\documentclass[conference]{IEEEtran}
\IEEEoverridecommandlockouts
\usepackage{cite}
\usepackage{amsmath,amssymb,amsfonts}
\usepackage{graphicx}
\usepackage{textcomp}

\usepackage[english]{babel}
\usepackage[utf8x]{inputenc}
\usepackage{caption}
\usepackage{subcaption}
\usepackage{epstopdf}

\def\BibTeX{{\rm B\kern-.05em{\sc i\kern-.025em b}\kern-.08em
    T\kern-.1667em\lower.7ex\hbox{E}\kern-.125emX}}
\begin{document}

\title{Representation Learning in Partially Observable Environments using Sensorimotor Prediction\\
}

\author{\IEEEauthorblockN{Thibaut Kulak}
\IEEEauthorblockA{\textit{SoftBank Robotics Europe - AI Lab} \\
thibaut.kulak@gmail.com}
\and
\IEEEauthorblockN{Michael Garcia Ortiz}
\IEEEauthorblockA{\textit{SoftBank Robotics Europe - AI Lab} \\
mgarciaortiz@softbankrobotics.com}
}

\maketitle

\begin{abstract}
In order to explore and act autonomously in an environment, an agent needs to learn from the sensorimotor information that is captured while acting. By extracting the regularities in this sensorimotor stream, it can learn a model of the world, which in turn can be used as a basis for action and exploration. 

This requires the acquisition of compact representations from a possibly high dimensional raw observation, which is noisy and ambiguous. In this paper, we learn sensory representations from sensorimotor prediction. We propose a model which integrates sensorimotor information over time, and project it in a sensory representation which is useful for prediction. We emphasize on a simple example the role of motor and memory for learning sensory representations. 
\end{abstract}

\begin{IEEEkeywords}
Representation Learning, Sensorimotor Prediction, Recurrent Neural Networks, Ambiguous Environments
\end{IEEEkeywords}

\section{Introduction}
Autonomous Learning for Robotics aims to endow (robotic) agents with the capability to learn from and act in their environment, so that it can adapt to previously unseen situations. 
In order to be able to learn from this interaction, an agent has to build compact representations of the environments, using information captured from a high dimensional raw input. 

Current approaches favor the learning of representations using Deep Neural Networks (\cite{Goodfellow-et-al-2016}, \cite{lecunreview}, \cite{DBLP:journals/corr/Schmidhuber14}). Supervised learning extracts representations from the data to solve a classification task, providing the agent with hierarchical compact representations of different sensory streams (~\cite{krizhevsky2012imagenet}, \cite{he2016deep}). However, these state-of-the-art machine learning algorithms are not suitable for autonomous learning, as they rely on labeled data, which are costly to acquire, and are constraining the representations on the classes they were trained on. 
Unsupervised learning allows to learn hierarchical compression for different data streams (\cite{hinton2006reducing}, \cite{tenenbaum2000global}, \cite{radford2015unsupervised}). These representations, based on the statistics of the data, are very efficient to reduce the dimensionality of the sensory stream. However, these representations fail to inform the agent on the modalities of its potential actions in its environment. This is related to the problem of grounding knowledge in the experience of an agent \cite{harnad}. 
More recently, Deep Reinforcement Learning proposed to learn sensory representations together with a policy to act in an environment \cite{mnih-dqn-2015} in order to solve a set of tasks. However, the learning  depends on the external reward provided to the agent, itself dependent on the task to solve. We don't use Reinforcement Learning because we don't want to influence the sensory representations with guided exploration.

Mainstream approaches to learning representations are not sufficient for open-ended learning.
Several works are exploring new theories to propose mechanisms for an autonomous agent to learn representations from its environment. In particular, sensorimotor prediction states that an agent learns the structure of its world by learning how to predict the consequences of its actions (\cite{o2001sensorimotor}, \cite{friston2010free}). This sensorimotor approach tries to bring together sensor representations and motor representations by identifying the regularities in the sensorimotor stream. 
However, these regularities are hard to capture: a robotic agent acts and perceives in an environment which is usually partially observable (limited field of view), noisy and ambiguous. The sensory information, or observation, is not sufficient to know the exact state of the agent in its environment (similar sensory states can originate from different situations in the environment). This is in particular true for navigation tasks where an agent can observe several occurrences of very similar portions of the scenes (wall, corners) at different locations in the environment ( e.g. in a maze).
For these reasons, we need representations that can help disambiguate the observations and the state of the agent. 

If we take inspiration from biology, animals, in particular rats, have the ability to localize and navigate in complex mazes, which constitute partially observable and ambiguous environments. They rely on high-level representations of their environment which they learn through experience: place cells and grid cells \cite{moser2015place}. Grid cells are encoding information about the ego-motion of the rat, and place cells are activated when the rats finds itself in very particular locations of the environments.
Even though many works have studied their role in cognition, the emergence of these cells are not yet completely understood. However, their nature tends to confirm that an agent should rely on a combination of sensory information, motor information and memory to help learning representations that it can use to infer its situation in an environment. 

In this paper, we propose to learn sensory representations using principles from sensorimotor prediction. We show that using motor information as well as a form of memory allows an agent to learn representations which lead to more accurate sensorimotor prediction, and that the representations learned are transferable across different environments. We show the benefits of this sensorimotor approach on a navigation scenario, where an agent equipped with distance sensors moves into an ambiguous maze.

\section{Related Work}
Our work is closely related to the literature about forward internal models \cite{wolpert1998internal}, which involve the learning of models to predict the sensory consequences of actions. For instance a forward model of physics is learned  for a real-world robotic platform in \cite{agrawal2016learning}. Nonetheless, our contribution is not to  build a forward internal model, but to show that doing so forces the emergence of good sensory representations. \\
Many works are involved with the learning of representations, especially with recent advances in Deep Learning and Representation Learning (\cite{Goodfellow-et-al-2016}, \cite{lecunreview}, \cite{DBLP:journals/corr/Schmidhuber14}). For conciseness, we focus on those that involve the emergence of representations from sensorimotor prediction in the case of an agent moving in a fixed environment.

In  \cite{agrawal2015learning}, the authors assume that a good image representation is one that allows to predict the ego-motion between two images. Siamese networks extract features from two images, and these features correspond to the sensory representations. They are used to predict the motor command corresponding to the change from one image to the other. This instance of sensorimotor learning shows that prediction can be used as a mechanism to build representations, but lacks a form of memory in the representation, which we believe is necessary to disambiguate between states that correspond to similar observations, but different localizations in the world.

In \cite{stachenfeld2017hippocampus}, the authors propose to learn a model of the environment of an agent by encoding the prediction of its future states into cells, which correspond to a form of sensorimotor representations. They showed that decomposing the state transition matrix leads to spatial representations akin to grid cells. This approach is very relevant, however they use the localization of the agent as input to extract higher level spatial representations, whereas in our case, these representations have to be built based on distance sensor data.

In \cite{antonelo2009unsupervised}, the authors show that slowness principle allows an agent equipped with laser sensors, moving in a maze, to develop representations akin to place cells. In \cite{schonfeld2015modeling}, the authors also propose to use slowness to extract spatial representations from visual input. In both cases, motor commands are not taken into account for representation learning, and the environment is simple compared to the dimensionality of the sensors, rendering the problem non ambiguous. Additionally, the environment is one-dimensional, and their approach is very dependent on a good exploratory behavior. For these reasons, we believe that the approach doesn't scale to ambiguous, multidimensional, partially observable environments. 


In \cite{jonschkowski2015learning}, the authors propose to learn a mapping from an RGB image to a representation space constrained by different priors (repeatability, proportionality with respect to the motor commands, causality and temporality), for a planar agent moving in a simple environment. They observe that the learned representation space corresponds to the position of the agent in the environment. This paper is very relevant, as the authors learn representations based on motor and sensors. However, the definition of their priors require having an external reward, which is not suitable for open-ended learning. Also, their environment is non-ambiguous, and the motor commands used are discrete very simple. 

We present a representation learning method based on sensorimotor prediction which allows an agent to make use of memory and sensorimotor streams to create representations of its environment. 

\section{Sensorimotor predictive model as representational learning framework}
Representation Learning methods are usually based on the minimization of the distance between an input and its reconstruction from the encoding of the input \cite{Goodfellow-et-al-2016}. We argue that reconstruction error alone doesn't capture important properties of the environment. Objects might be very close in $\mathcal{L}_2$ distance but very different in nature. 
For example, an agent perceiving with range sensors will consider that a wall and a corner at the same distance from the agent are very similar, whereas 2 walls at different distances from the agents might be considered as very dissimilar. We propose to learn representations based on a prior inspired from sensorimotor theories of perception (\cite{o2001sensorimotor}, \cite{friston2010free} ) : an agent acquires knowledge about its environment by building a sensorimotor predictive model. This model takes as inputs raw sensor and motor values.

\subsection{Recurrent Sensorimotor Encoder}
\begin{figure}[h!]
\centering
\includegraphics[width=0.4\textwidth]{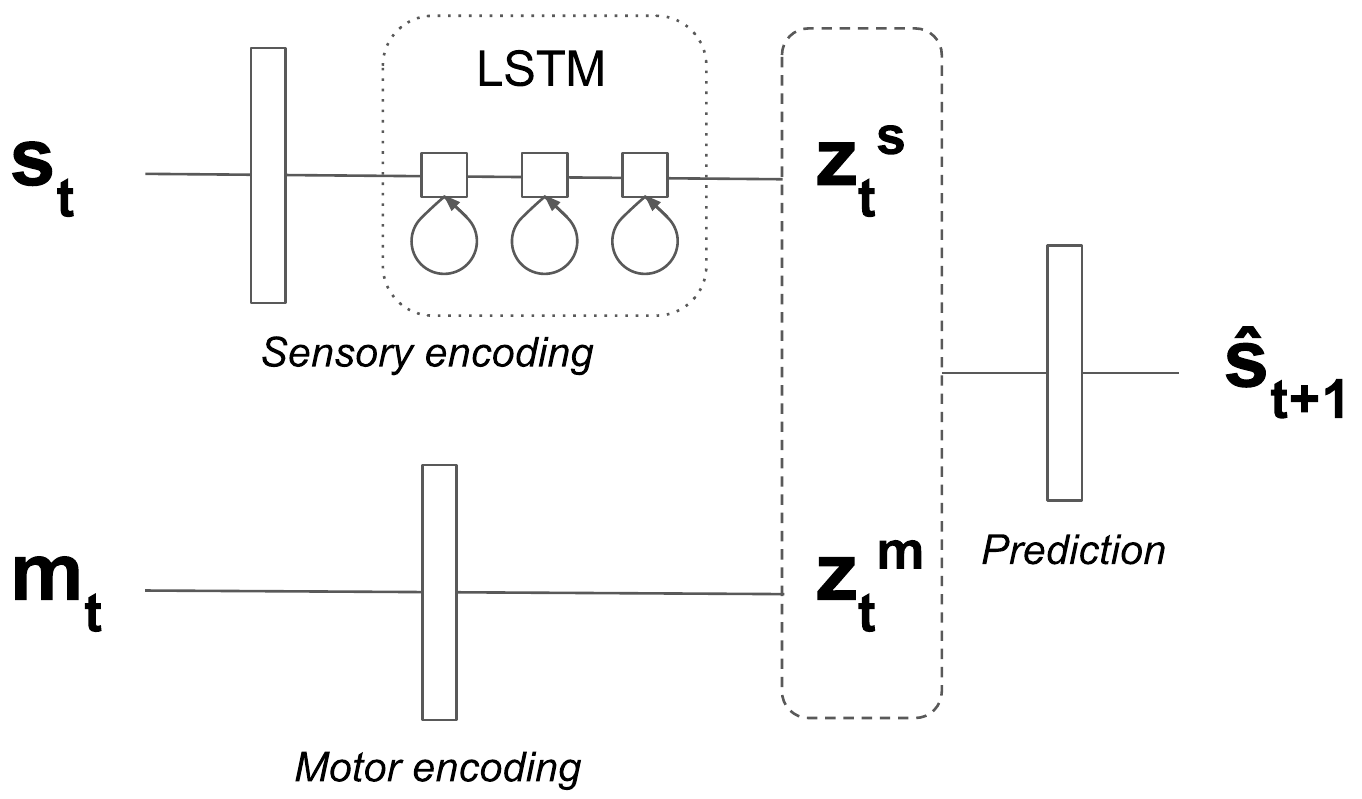}
\caption{Recurrent Sensorimotor Encoder}
\label{fig:lstmsm_architecture}
\end{figure}

We propose a model named Recurrent Sensorimotor Encoder (abbreviated Recurrent-SM-encoder), for representation learning using sensorimotor prediction and memory. 
It is shown in Fig.\ref{fig:lstmsm_architecture} and it is composed of three subnetworks : \\
(i) A sensory encoding subnetwork takes as input the sensory state $s_{t}$ and outputs an encoded sensory state $z_{t}^{s}$. It is composed of hidden layers followed by a stacked LSTM network, which role is to provide a form of memory about the previous sensor states.
(ii) A motor encoding subnetwork, which is a classical dense network composed of hidden layers, taking as input the motor command $m_{t}$ and outputting the encoded motor command $z_{t}^{m}$.
(iii)  $z_{t}^{s}$ and $z_{t}^{m}$ are concatenated to form the encoded sensorimotor vector $z_{t}^{sm}$, used as an input for a dense network, which outputs a prediction of the next sensory state $\hat{s}_{t+1}$

This deep network is trained end-to-end with a sensorimotor prediction loss on the output space (see Eq.\ref{eq:loss}).

\begin{equation}
Loss \left( (s_{t},m_{t})_{t=1}^{T} \right) = \sum_{t=1}^{T-1} \left( \hat{s}_{t+1} - s_{t+1} \right) ^2
\label{eq:loss}
\end{equation}

We chose to separate sensory and motor representations, because ultimately these representations might be shared to solve different tasks.

\subsection{Baselines}
\begin{figure}[h!]
\noindent
\centering

\begin{minipage}[t]{0.3\textwidth}
\centering
\includegraphics[width=0.8\textwidth]{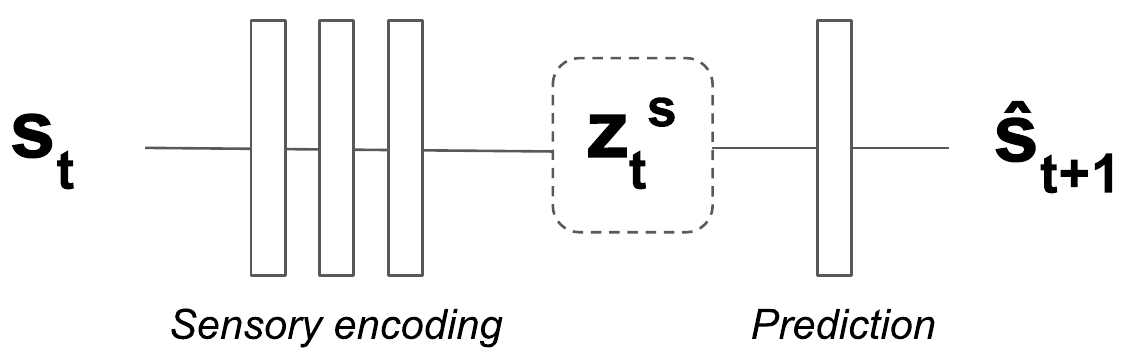}
\caption*{(a) Sensory Encoder}
\label{fig:encoders_architecture}
\end{minipage}

\vspace{\baselineskip}

\begin{minipage}[t]{0.37\textwidth}
\centering
\includegraphics[width=0.8\textwidth]{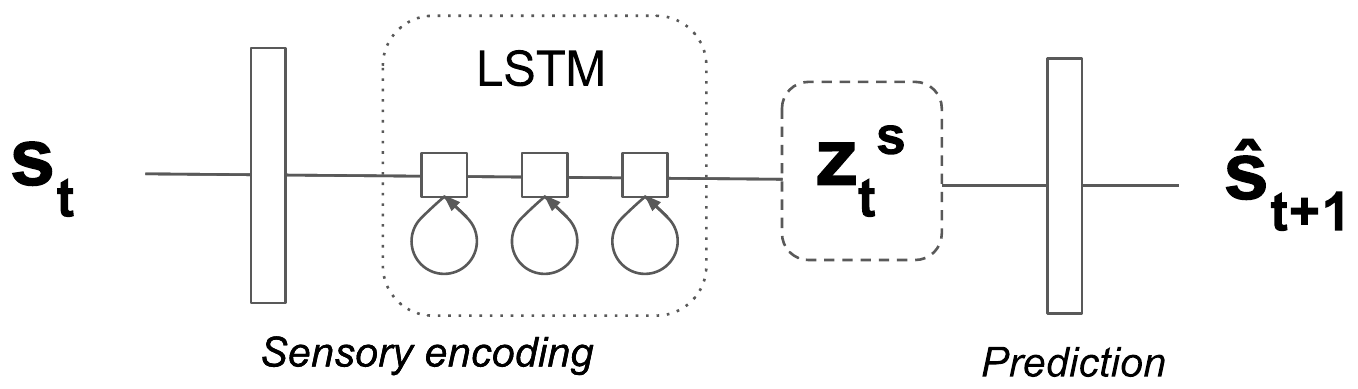}
\caption*{(b) Recurrent Sensory Encoder}
\label{fig:lstms_architecture}
\end{minipage}

\vspace{\baselineskip}

\begin{minipage}[t]{0.32\textwidth}
\centering
\includegraphics[width=0.8\textwidth]{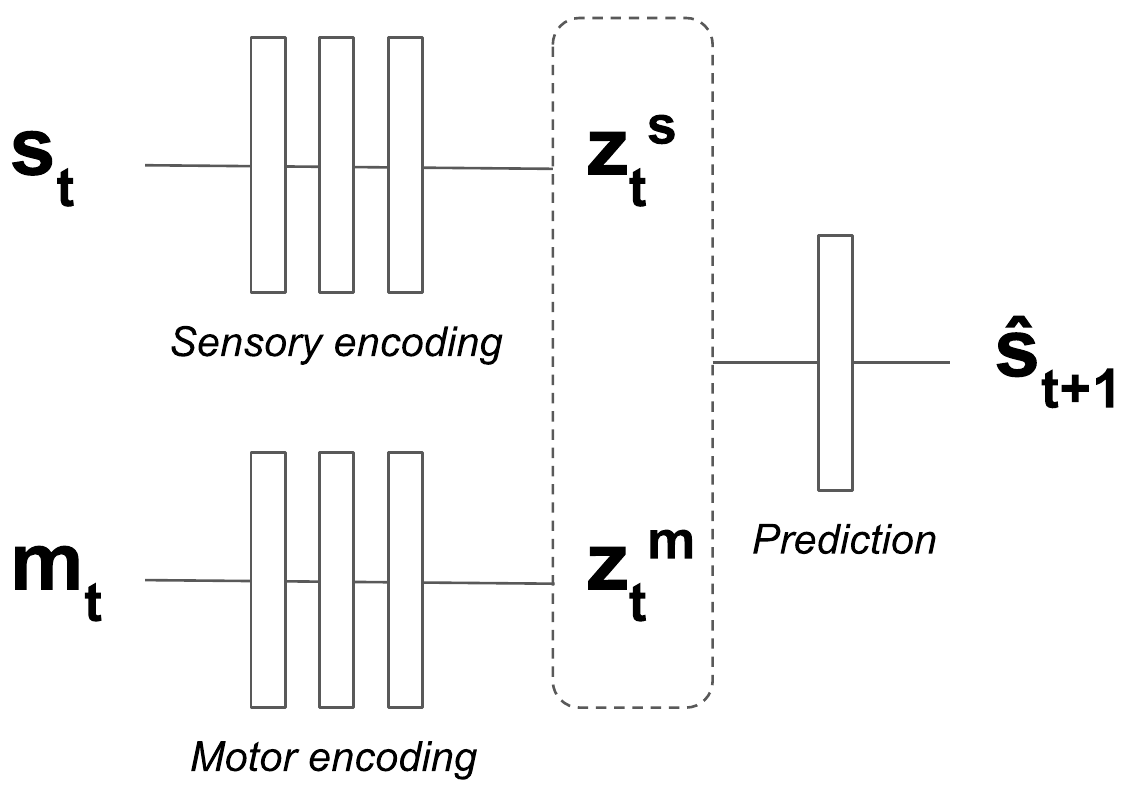}
\caption*{(c) Sensorimotor Encoder}
\label{fig:encodersm_architecture}
\end{minipage}

\caption{Architectures of the baselines}
\label{fig:baselines_architectures}
\end{figure}

In order to evaluate the effect of sensorimotor prediction and the use of a memory (through LSTM) on the quality of the sensory representation learned, we introduce several baselines to compare our system. 
The first one, called Sensorimotor Encoder (abbreviated SM-encoder), doesn't have a memory. The second one, named Recurrent Sensory Encoder (abbreviated Recurrent-S-encoder), doesn't take into account the motors. The third baseline, the Sensory Encoder (abbreviated S-encoder), encodes the raw sensory input, without memory or motors.
These baselines are presented in Fig.\ref{fig:baselines_architectures}.

\section{Experimental setup}
We developed a simulation of an agent equipped with laser sensors in partially observable environments, in order to build a database with sequences of sensory states and motor commands. The simulation is simple enough so that it can be easily reproduced.

\subsection{Agent}
Our simulated agent is loosely inspired from the Thymio-II robot \cite{riedo2012two}. It is equipped with 5 distance sensors, evenly separated between -0.6 and 0.6 radians, and their range is limited to 10 units of distance. 
The agent has one motor command controlling the translation forward (direction of the middle laser), the second one controls the rotation. One motor command $(d,r)$ is the succession of a translation $d$ and a rotation $r$. 
It is a planar agent moving without friction, and there is no noise on its distance sensors. 


\subsection{Environments}
\begin{figure}[h]
\centering

\begin{minipage}{0.125\textwidth}
\centering
\includegraphics[width=\textwidth]{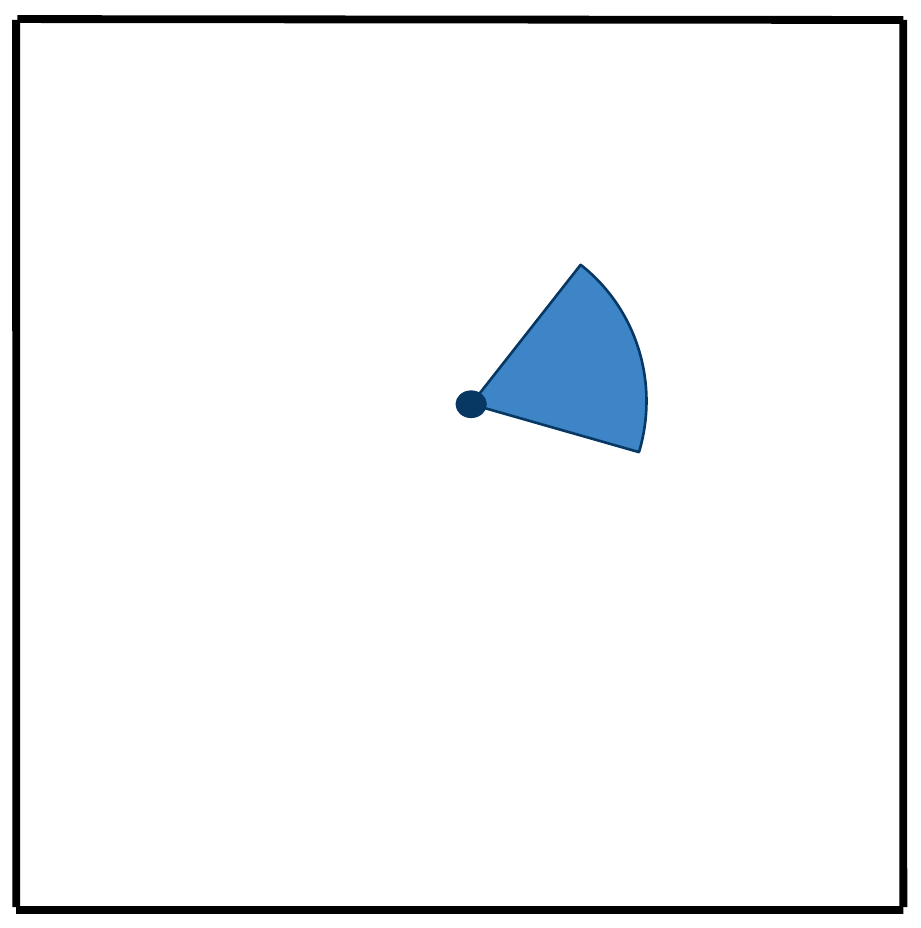}
\\ (a) Square
\label{fig:env1}
\end{minipage}
\hspace{\stretch{1}}
\begin{minipage}{0.125\textwidth}
\centering
\includegraphics[width=\textwidth]{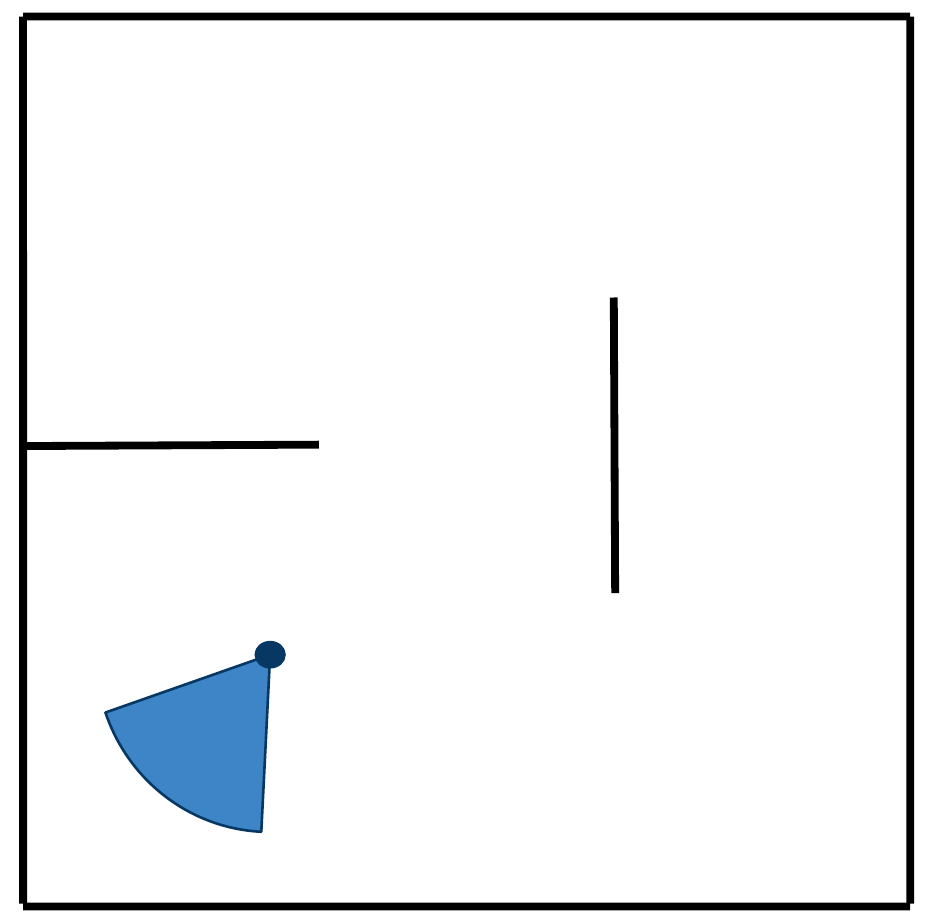}
\\ (b) Rooms1
\label{fig:env2}
\end{minipage}
\hspace{\stretch{1}}
\begin{minipage}{0.125\textwidth}
\centering
\includegraphics[width=\textwidth]{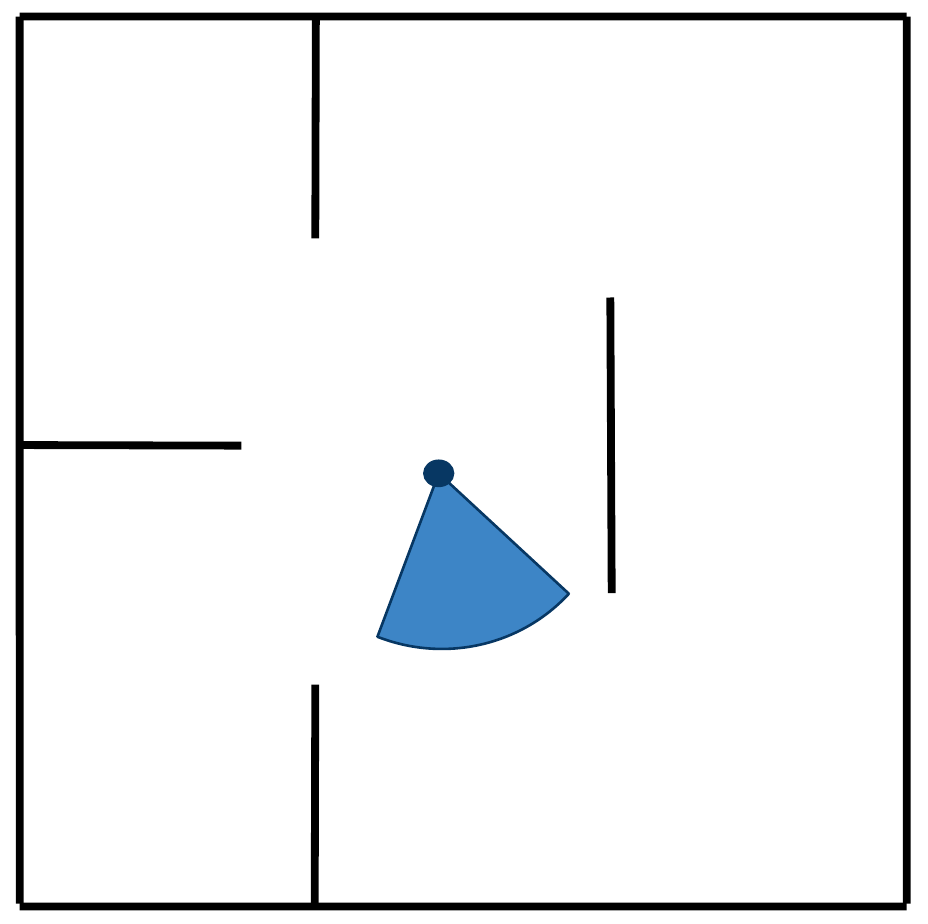}
\\ (b) Rooms2
\label{fig:env3}
\end{minipage}

\caption{The different environments created.}
\label{fig:environments}
\end{figure}

We created 3 environments of size 50 units, shown on Fig.\ref{fig:environments}.
The first environment, named Square, is a square without walls or obstacles. The second environment, named Rooms1, contains one vertical wall and one horizontal wall.
The second environment, named Rooms2, contains one horizontal and three vertical walls.

\subsection{Behavior}
\label{sec:behavior}
The agent moves by a random translation forward and a random small rotation at each time-step. To avoid collisions with the walls, when one of its distance sensors gets below a threshold, it turns around.
Formally, at each timestep, if one distance sensor value is smaller than 1 unit, the agent rotates by $r \sim\ \mathbf{U}(\pi - \frac{\pi}{10}, \pi + \frac{\pi}{10})$ radians ($\mathbf{U}$ denoting the uniform distribution.) If not, the agent moves forward by $ d \sim\ \mathbf{U}(0,1)$ units, and rotates by $r \sim\ \mathbf{U}(-\frac{\pi}{6},\frac{\pi}{6})$ radians. This allows to simplify the sensorimotor modalities we are studying, by preventing collisions during the exploration.

\subsection{Databases}
\begin{figure}[h!]
\noindent

\begin{subfigure}{0.15\textwidth}
\centering
\includegraphics[width=\textwidth]{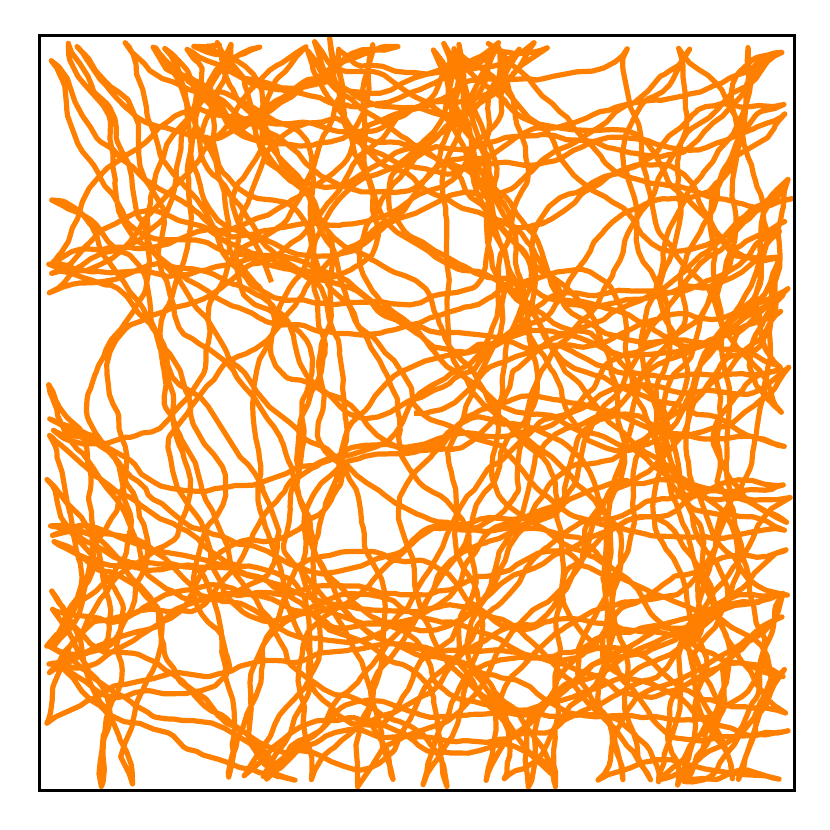}
\captionof{figure}{Square}
\label{fig:trajectory_square}
\end{subfigure}
\hspace{\stretch{1}}
\begin{subfigure}{0.15\textwidth}
\centering
\includegraphics[width=\textwidth]{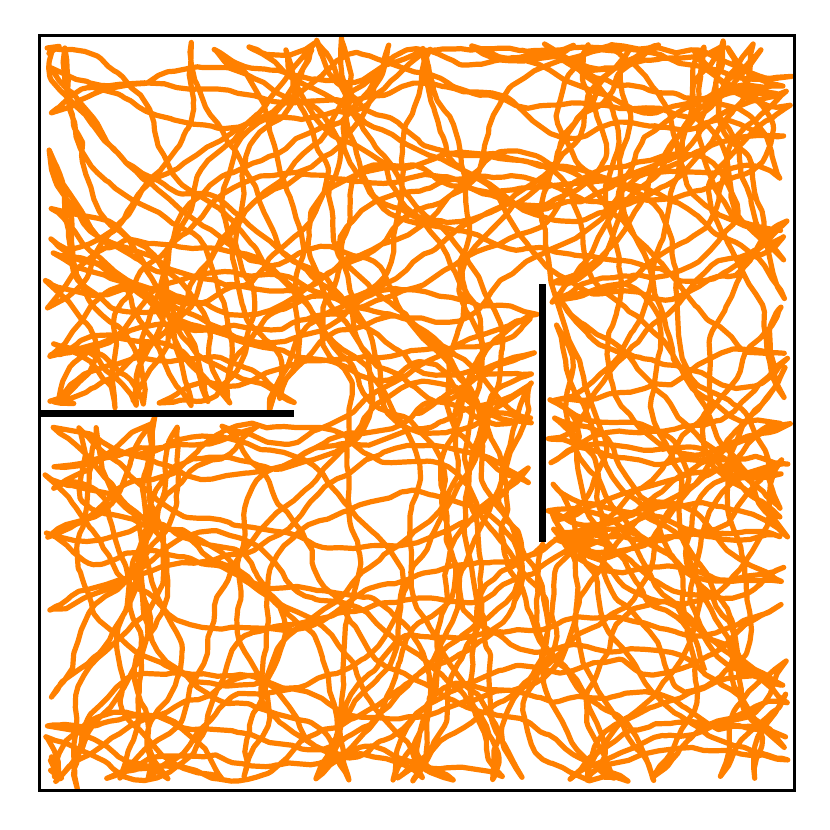}
\captionof{figure}{Rooms1}
\label{fig:trajectory_env2}
\end{subfigure}
\hspace{\stretch{1}}
\begin{subfigure}{0.15\textwidth}
\centering
\includegraphics[width=\textwidth]{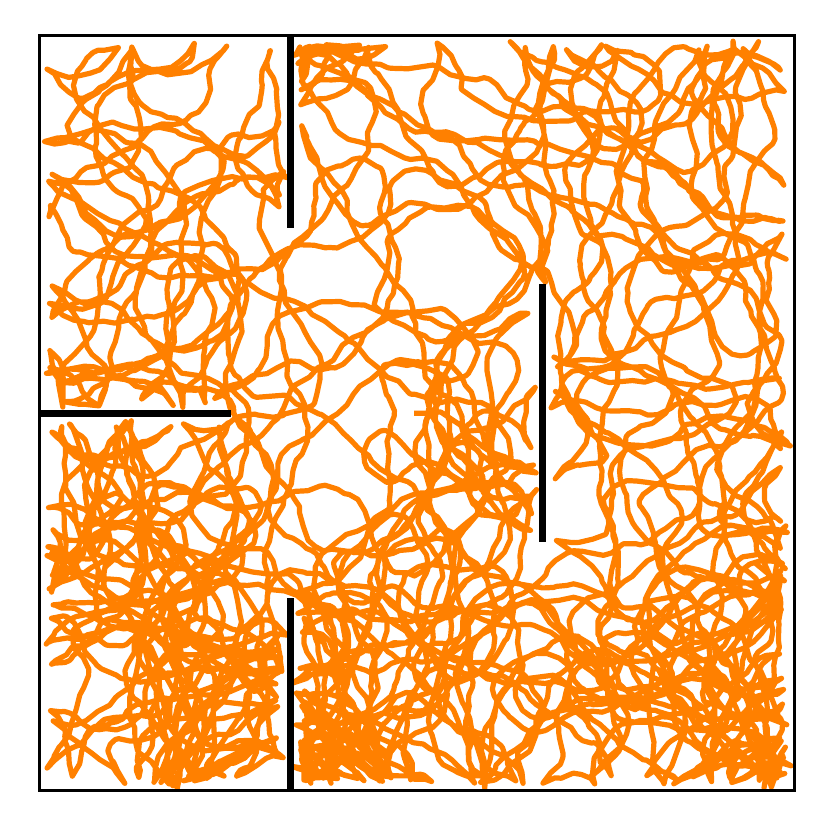}
\captionof{figure}{Rooms2}
\label{fig:trajectory_env3}
\end{subfigure}
\caption{Trajectories in the environments (10 000 points)}
\label{fig:trajectories}
\end{figure}

\begin{figure*}[h!]
\noindent

\begin{minipage}[t]{0.15\textwidth}
\centering
\includegraphics[width=\textwidth]{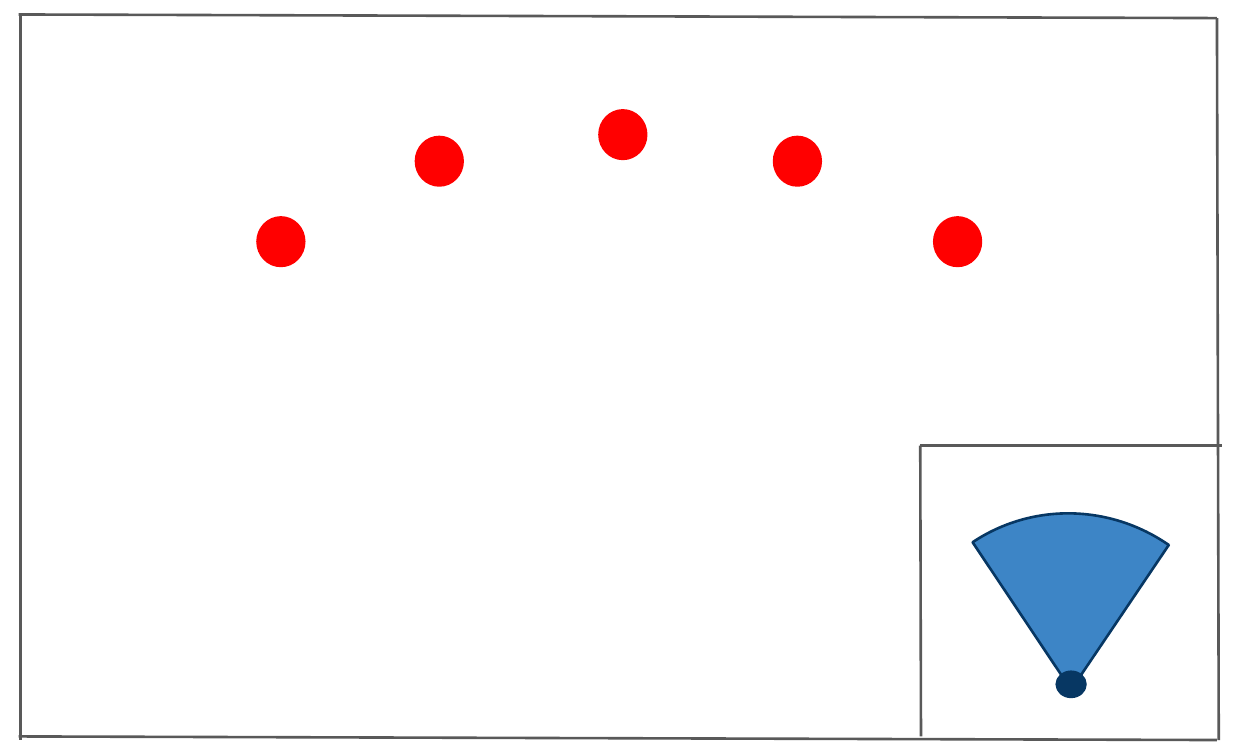}
\\ (a) Agent perceiving nothing
\label{fig:database_screenshot_nothing}
\end{minipage}
\hspace{\stretch{1}}
\begin{minipage}[t]{0.15\textwidth}
\centering
\includegraphics[width=\textwidth]{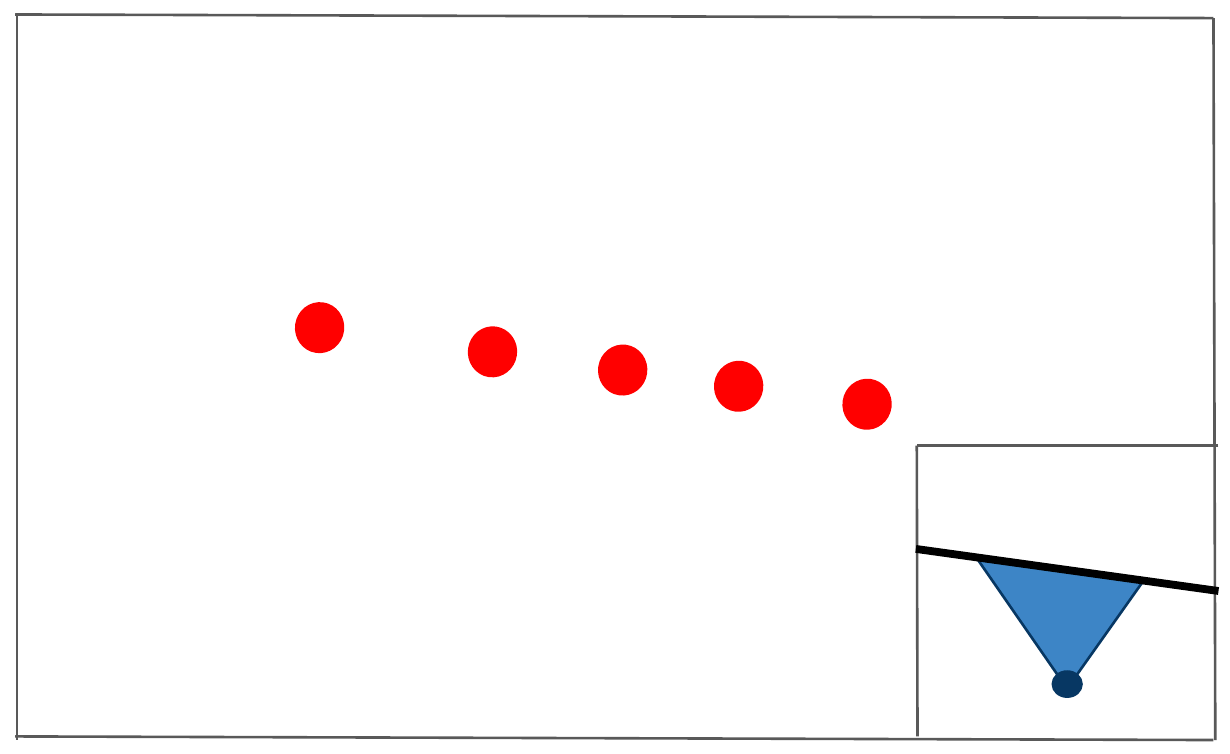}
\\ (b)  Agent in front of a line
\label{fig:database_screenshot_line}
\end{minipage}
\hspace{\stretch{1}}
\begin{minipage}[t]{0.15\textwidth}
\centering
\includegraphics[width=\textwidth]{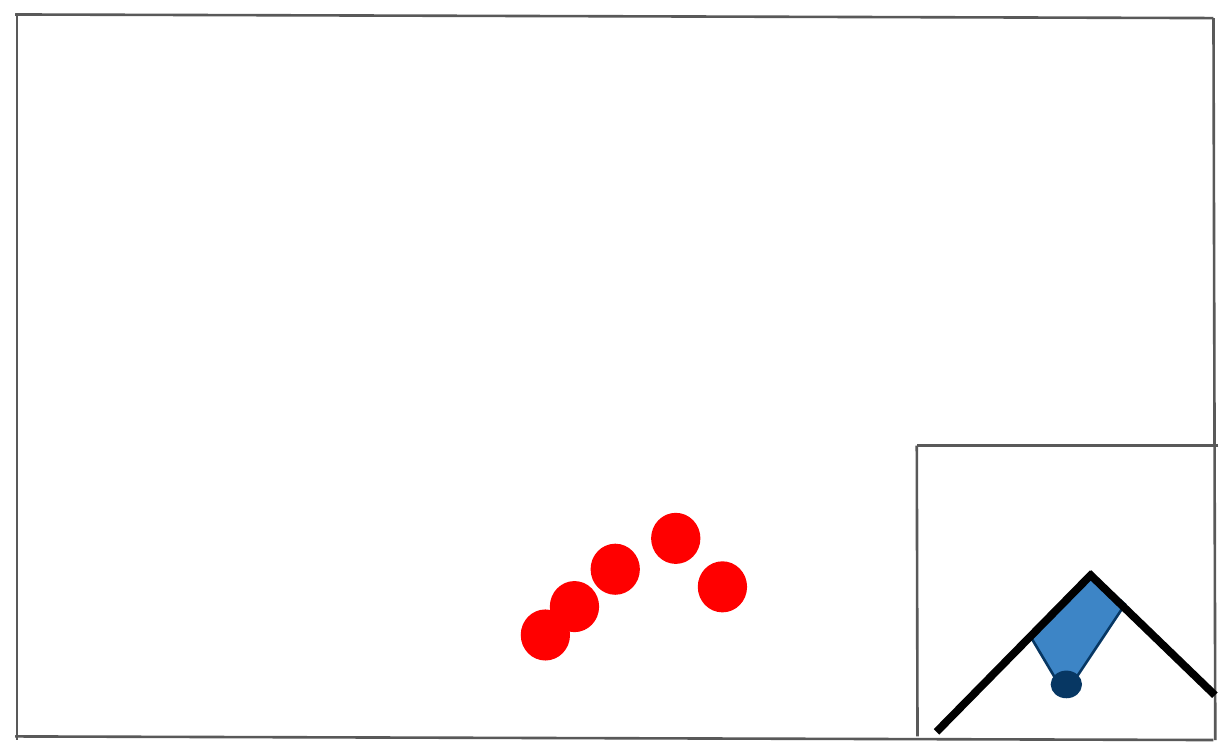}
\\ (c) Agent in front of a corner
\label{fig:database_screenshot_corner}
\end{minipage}
\hspace{\stretch{1}}
\begin{minipage}[t]{0.15\textwidth}
\centering
\includegraphics[width=\textwidth]{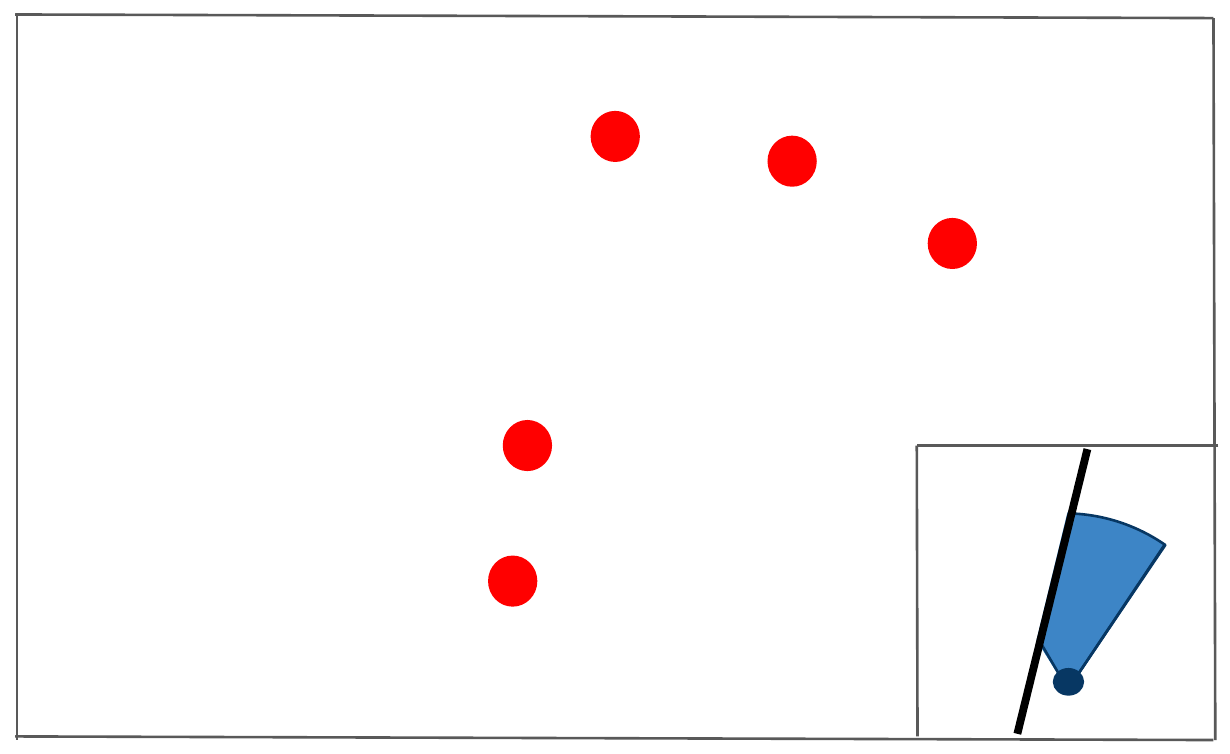}
\\ (d) Agent with a wall at its left
\label{fig:database_screenshot_wallleft}
\end{minipage}
\hspace{\stretch{1}}
\begin{minipage}[t]{0.15\textwidth}
\centering
\includegraphics[width=\textwidth]{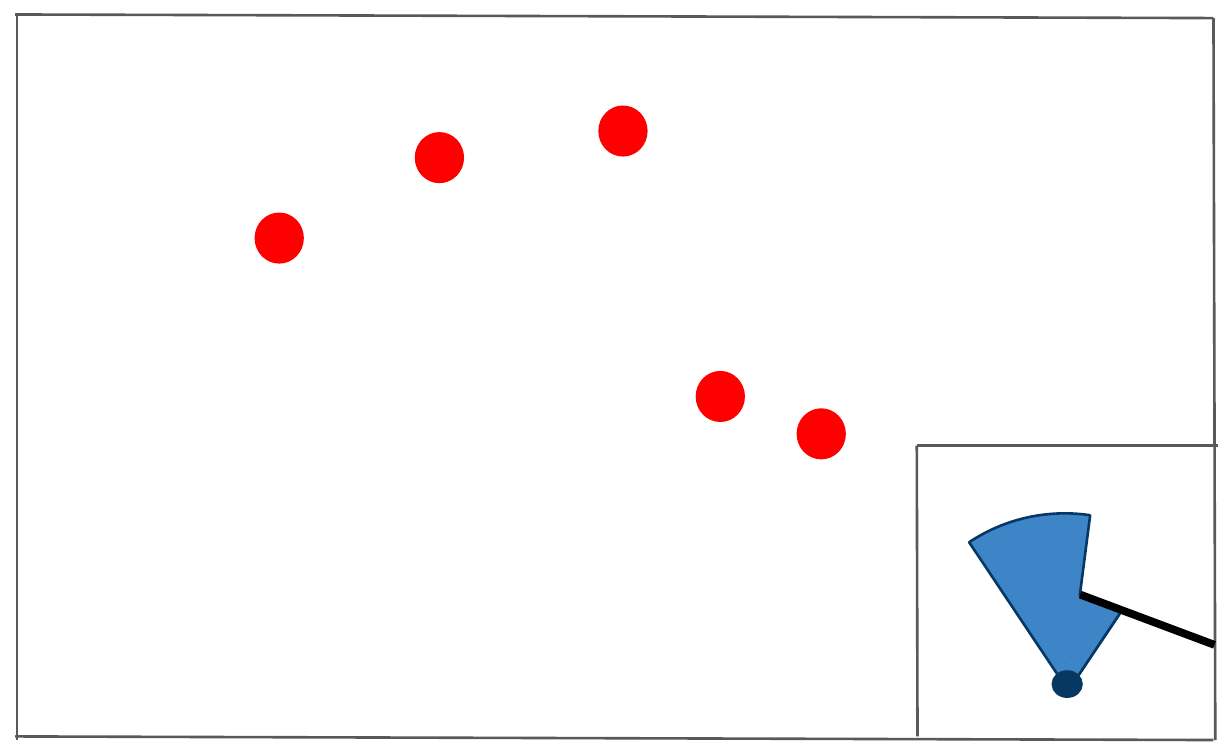}
\\ (e) Agent in front of a wall's end
\label{fig:database_screenshot_together}
\end{minipage}

\caption{Examples from the databases. The 5 red dots represent the distance perceived by the agent, projected in top-view.}
\label{fig:database_screenshots}
\end{figure*}

Our databases are generated sequences of 1 000 000 points (each point has 5 distance sensors values and 2 motor commands), which we have separated as such : the first 80\% for training, the next 10\% for validation, and the last 10\% for testing. In Fig.\ref{fig:database_screenshots}  we reconstructed, for different situations, what the agent perceives based on its sensors. Note that the agent doesn't have access to the position and angles of its distance sensors, it only receives as input a 5-dimensional real vector. Fig.\ref{fig:trajectories} displays the trajectory of the agent during 10 000 steps in the different proposed environments: Square, Rooms1 and Rooms2. We can see that the chosen behavior allows a complete  exploration of each environment.

\begin{figure*}[h!]
\centering
\noindent

\begin{subfigure}[t]{.15\textwidth}
\centering
\includegraphics[height=2.5cm]{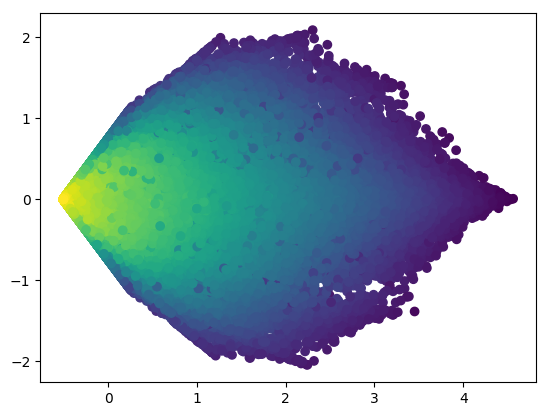}
\caption{\centering Sensory}
\label{fig:sensoryspace}
\end{subfigure}
\hspace{\stretch{10}}
\begin{subfigure}[t]{.15\textwidth}
\centering
\includegraphics[height=2.5cm]{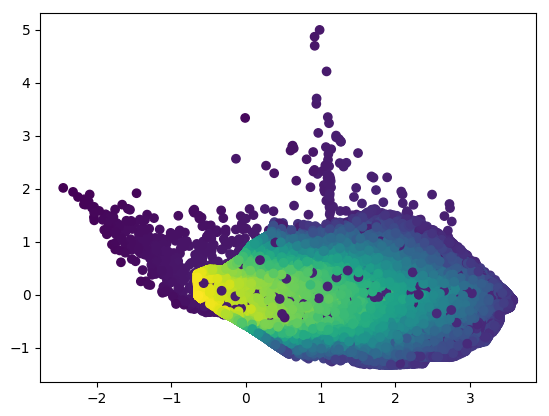}
\caption{\centering Sensory encoder}
\label{fig:encoder_sm_nomotors}
\end{subfigure}
\hspace{\stretch{10}}
\begin{subfigure}[t]{.15\textwidth}
\centering
\includegraphics[height=2.5cm]{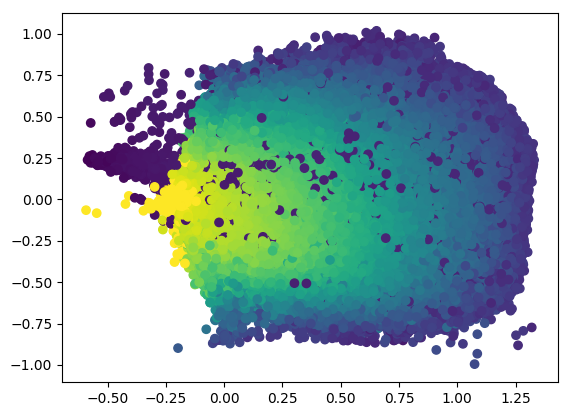}
\caption{\centering Recurrent sensory encoder}
\label{fig:lstm_sm_nomotors}
\end{subfigure}
\label{fig:comparisonRepresentationSpacesSquare}
\hspace{\stretch{10}}
\begin{subfigure}[t]{.15\textwidth}
\centering
\includegraphics[height=2.5cm]{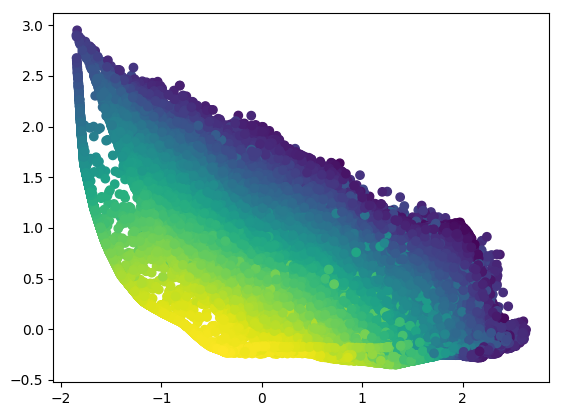}
\caption{\centering Sensorimotor encoder}
\label{fig:encoder_sm}
\end{subfigure}
\hspace{\stretch{10}}
\begin{subfigure}[t]{.15\textwidth}
\centering
\includegraphics[height=2.5cm]{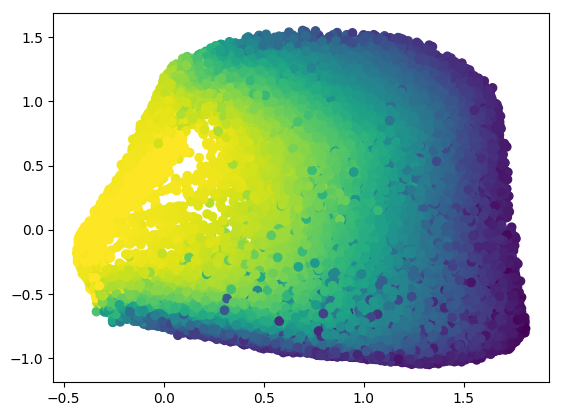}
\caption{\centering Recurrent sensorimotor encoder}
\label{fig:lstm_sm}
\end{subfigure}
\hspace{\stretch{10}}
\begin{subfigure}[t]{.05\textwidth}
\centering
\includegraphics[height=2.5cm]{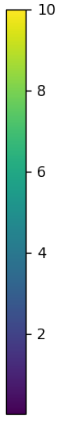}
\end{subfigure}
\caption{Representation spaces learned on Square, colored by the minimum value of the lasers}
\label{fig:comparisonRepresentationSpacesSquare}
\end{figure*}

\section{Results}
\subsection{Numerics}
Our models are trained with the Adam optimizer (learning rate of 0.001) and early stopping. The training is stopped if the loss on the validation set doesn't decrease by 5\% for 10 consecutive epochs. We use a batch size of 64, and ReLUs for the activation functions. We choose arbitrarily the sensory representation space to be 10-dimensional and the motor representation space to be 5-dimensional. The number and size of layers in the different architectures are as follow:
\subsubsection{SM-encoder} The sensory encoding and motor encoding subnetworks have 3 hidden layers of size 16, 32 and 64, while the prediction subnetwork has one layer of size 128.
\subsubsection{S-encoder} It is identical to the the sensory encoder, without the motor encoding subnetwork.
\subsubsection{Recurrent-SM-encoder} The sensory encoding and motor encoding subnetworks have 1 hidden layer of size 16, while the prediction subnetwork has one layer of size 128. The (stacked) LSTM has 3 layers with 32 units at each layer, and a truncation horizon of 20.
\subsubsection{Recurrent-S-encoder} It is identical to the Recurrent sensorimotor encoder, without the motor encoding subnetwork.

\subsection{Sensorimotor prediction results}
We report in Tab.\ref{table:square_results} the sensorimotor prediction $\mathcal{L}_2$ error of the models trained on
the Square environment, and tested on the three environments. We verified that the errors when training on Rooms1 and Rooms2 are comparatively the same for the different models, although obviously the error of each model increases as the environment becomes more difficult. For the sake of conciseness, we don't report them. We observe that models with motors largely outperform those without, which makes sense because motors are necessary to predict the next sensory state. We also see that models using a memory (with the LSTM) perform better compared to their memoryless counterpart, suggesting that a memory is useful for accurate sensorimotor prediction. Finally, we note that the Recurrent-SM-encoder model performs best, not only on the environment it is trained on, but also when tested on other environments, suggesting that the benefits of the memory provided by the LSTM isn't specific to the training environment.

\begin{table}[h!]
\centering
\begin{tabular}{| l | c | c |c |}
\hline
  Model & Square & Rooms1 & Rooms2 \\
\hline
  S-encoder & 0.03735 & 0.04299 & 0.07290 \\
  SM-encoder & 0.005576 & 0.01453 &  0.02572\\
  Recurrent-S-encoder &  0.03587 & 0.04068 & 0.06971\\
  Recurrent-SM-encoder & \textbf{0.002404} & \textbf{0.01054} & \textbf{0.01809}\\
\hline
\end{tabular}
\caption{Sensorimotor prediction $\mathcal{L}_2$ error of the models trained on Square tested on the test dataset of the three environments}
\label{table:square_results}
\end{table}

\subsection{Representation spaces}

In this illustrative section we propose to gain insight on what the sensory representations learned by our models represent.
We plot on Fig.\ref{fig:comparisonRepresentationSpacesSquare} the representation spaces learned by our models. Because those spaces are 10-dimensional we plot the 2D projection to the first two principal components extracted with PCA. For visualization purposes we color those spaces by the minimum value of the 5 lasers, because this gives information about the distance to the wall the agent perceives.

We observe that the models without motors group states where the agent doesn't see anything with states where the agent sees a wall from a very short distance. This makes sense, as the agent turns around when it is too close from a wall (\ref{sec:behavior}), and therefore goes with just one transition from seeing a wall very close to seeing nothing. Without access to motor commands, the model brings those states close to each other, while in reality those states are fundamentally different.

We see that the portion of the representation space corresponding to the agent perceiving nothing is bigger with the Recurrent-SM-encoder than with the Recurrent-S-encoder. This suggests that memory and the information about motor commands help creating different states for points where the agent doesn't see anything.

\begin{figure*}[h!]
\centering
   \includegraphics[width=\textwidth]{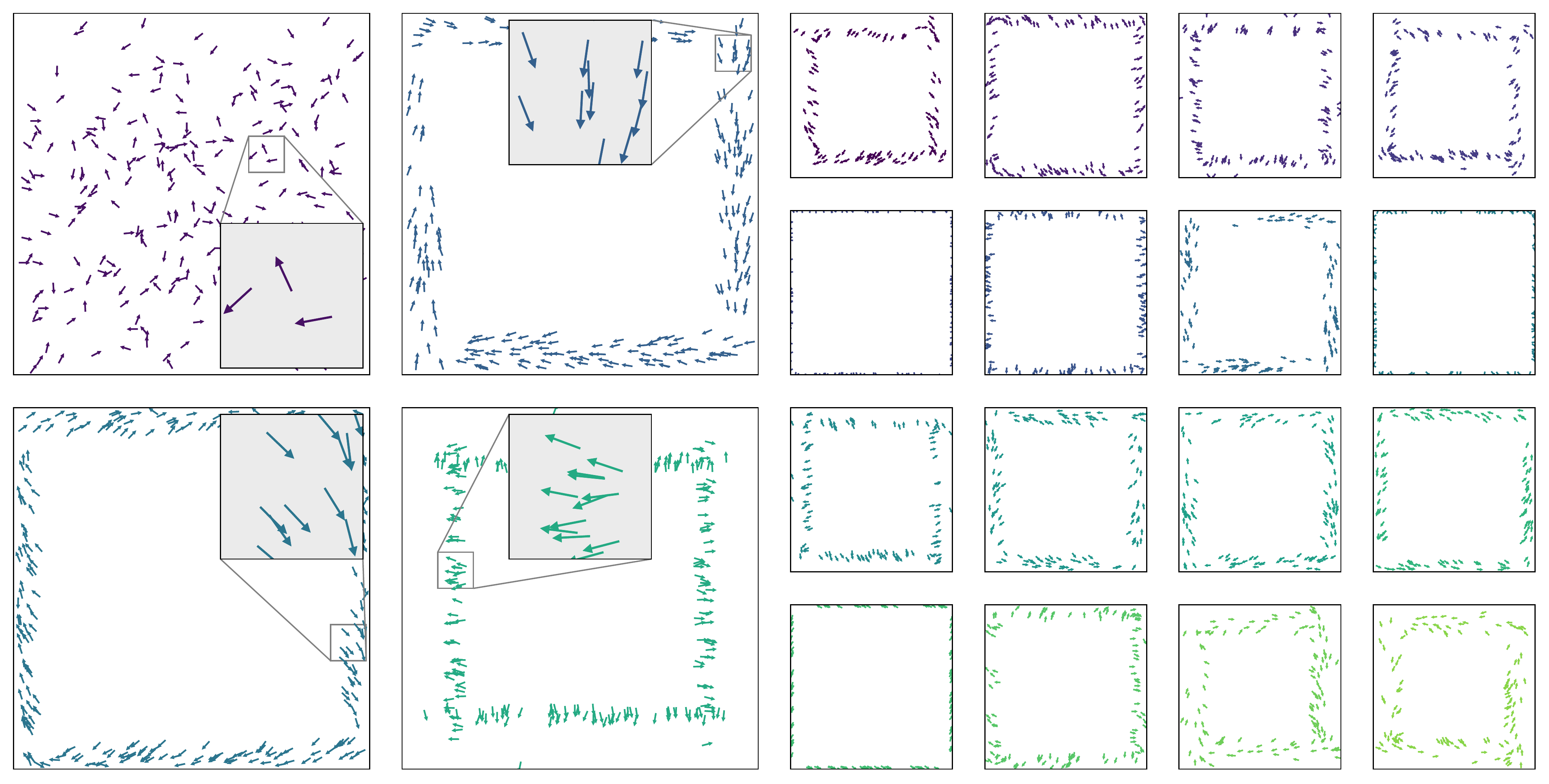}
   \caption{S-encoder representation space clusters}
   \label{fig:encoderS_square_square}
\end{figure*}

Interpreting qualitatively the distribution of the representation space gives us hints about the behavior of the models, and what they learn. In the following section, we propose other qualitative evaluations of our proposed systems.

\subsection{Clusters extraction}

We propose to cluster the sensory representation spaces, and to visualize the activation of the different clusters in space, in order to estimate if the sensory encoding learns spatial features. The clustering is done only for visualization purposes, in order to gain further insight on what the representations learned contain. We use a kMeans algorithm on the Sensory Representations to extract 20 clusters from each representation space.
We plot on separate subplots for each cluster the ground truth position and orientation of 500 data points of this cluster randomly sampled from the whole cluster data points. 

We show on Fig.\ref{fig:encoderS_square_square} the 20 clusters extracted from the S-encoder representation space. We see that there are clusters corresponding to different distances/angles to the wall. As there is no memory in this model all of the configurations when the agent doesn't perceive anything are in the same cluster. The clusters are on the whole the same as those obtained when clustering directly the sensory space, for that reason we don't show the latter.

\begin{figure*}[h!]
\centering
   \includegraphics[width=\textwidth]{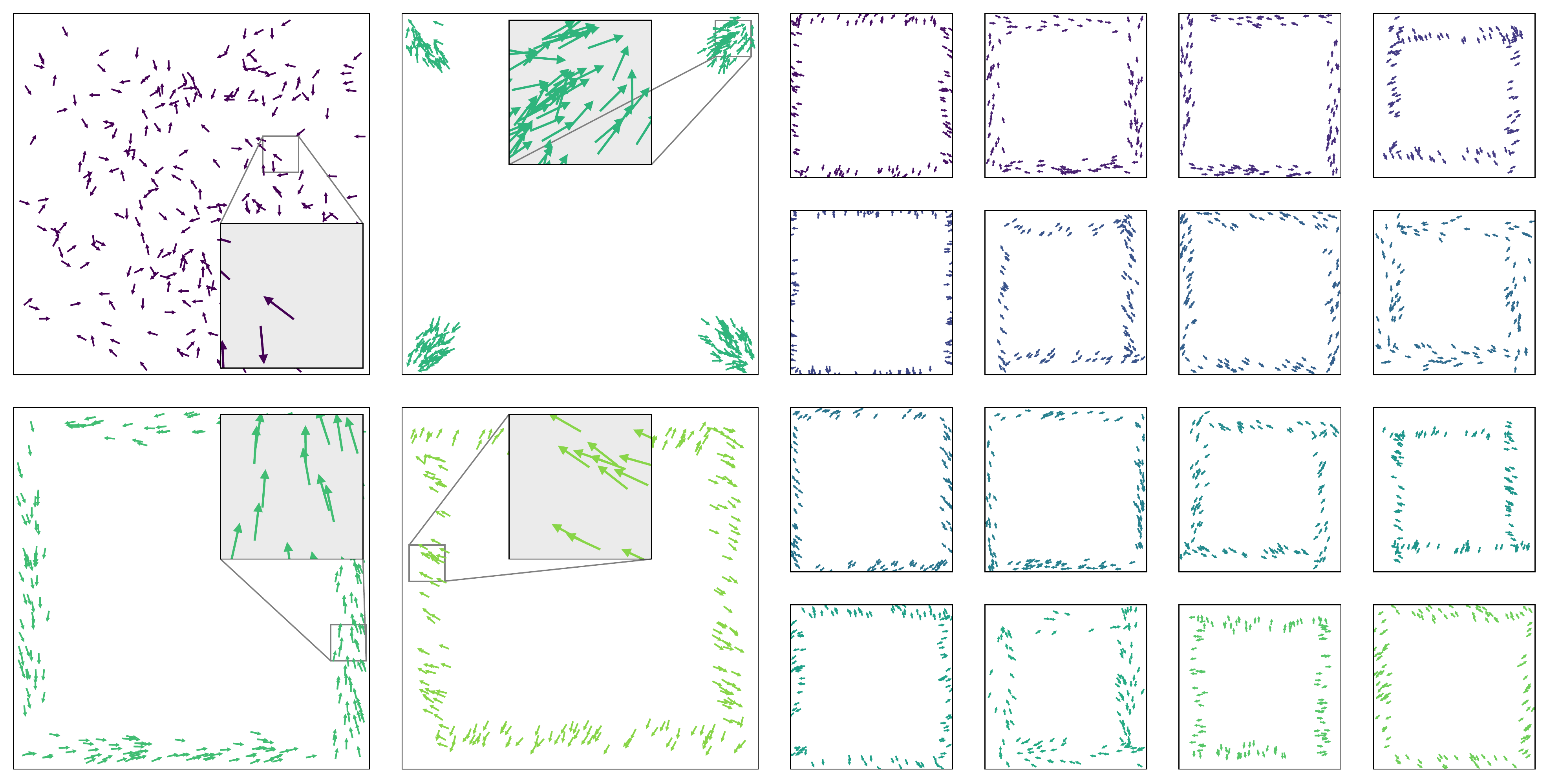}
   \caption{SM-encoder representation space clusters}
   \label{fig:encoderSM_square_square}
\end{figure*}

We observe on Fig.\ref{fig:encoderSM_square_square}, by comparing S-encoder and SM-encoder, that using motors allows the emergence of a cluster corresponding to corners of the environment. This is remarkable because the agent observes few corners (less than 1\% of the database).  It suggests that sensorimotor prediction is a good prior for representation learning because it might allow to create representation for rare occurances, which are different  from the sensorimotor point of view, i.e. if the interaction with them is different. This wouldn't be possible with a reconstruction loss, because a corner and a wall seen from the same distance are too close in $\mathcal{L}_2$ norm. We actually observed empirically that clustering the sensory space directly didn't permit the emergence of the cluster corresponding to corners.

\begin{figure*}[h!]
\centering
   \includegraphics[width=\textwidth]{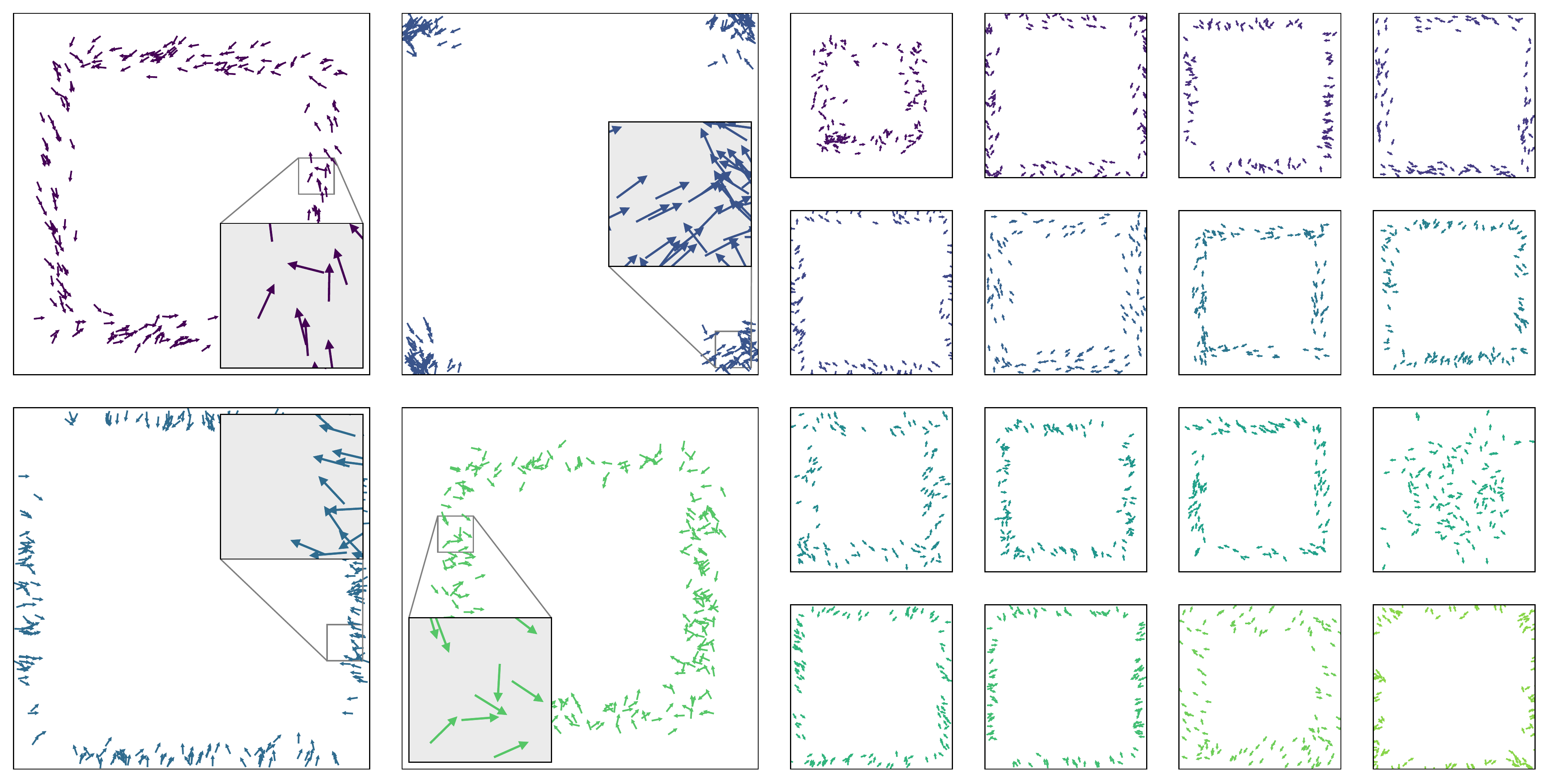}
   \caption{Recurrent-SM-encoder representation space clusters}
   \label{fig:lstmSM_square_square}
\end{figure*}

We see on Fig.\ref{fig:lstmSM_square_square} that the clusters extracted from the Recurrent-SM-encoder representation space also contain a cluster corresponding to corners. There are also clusters corresponding to different distances/angles to walls. We observe that we have different clusters corresponding to the agent perceiving nothing. These clusters correspond for instance to having a wall just behind, to having a wall behind at distance 10, and having a wall behind at the left at distance 5. This shows that the LSTM provides the agent with a memory of previous events, and that it contains a form of spatial information. However this memory is short-term as it is relative to the previous wall that has been seen, and there is no global notion of position in the environment.

We verified that when extracting clusters from the Recurrent-S-encoder representation space, no cluster corresponding to corners emerges, and we observed that there is also some memory but it is much less accurate than the one of the same model with motors. For the sake of conciseness, we decide not to show them.

\subsection{Robustness to training environment}

\begin{figure*}[h!]
\centering
   \includegraphics[width=\textwidth]{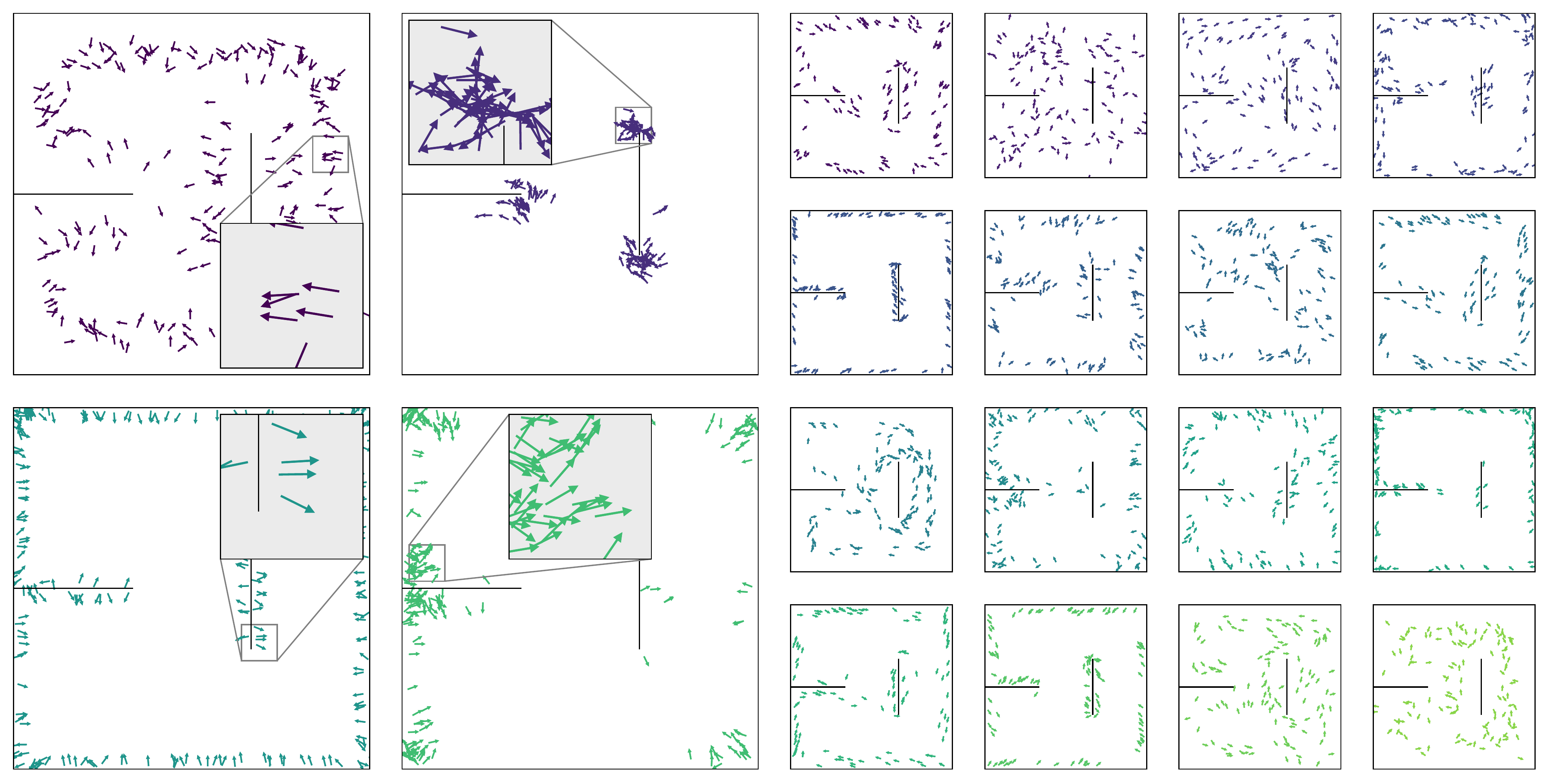}
   \caption{Recurrent sensorimotor encoder representation space clusters, trained on Rooms1}
   \label{fig:lstmSM_env2_env2}
\end{figure*}

In this experiment we evaluate the robustness of our approach when training on a more complex environment: Rooms1. We show on Fig.\ref{fig:lstmSM_env2_env2} the clusters extracted from the Recurrent-SM-encoder model trained on this environment. We observe that in addition to clusters similar to those appearing in Square environment, there is now a cluster corresponding to wall's ends. This shows that our model is robust to the training environment as it identified in this new environment that there is another interaction possibility with the environment, corresponding to wall's ends.

We saw however that when training on Rooms2, the latter cluster corresponding to wall's ends didn't emerge. We hypothesize that this is due to the fact that Rooms2 contains more walls, causing the agent's blocking in the different rooms for a long period of time, and therefore making the appearance of wall's ends in the database occasional.

\subsection{Robustness to testing environment}
\begin{figure}
\centering
\noindent

\begin{subfigure}[t]{.15\textwidth}
\centering
\includegraphics[width=\textwidth]{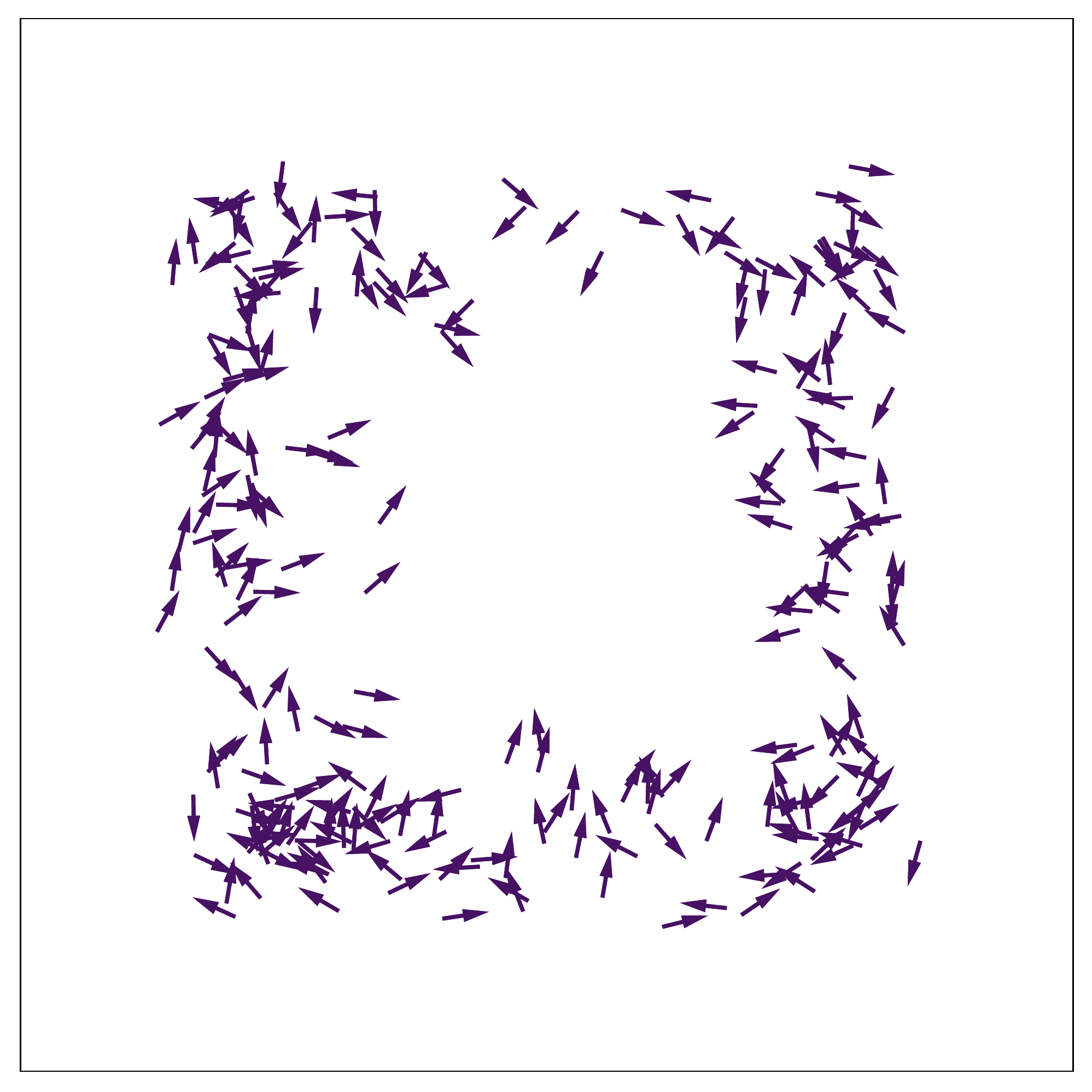}
\end{subfigure}
\hspace{\stretch{1}}
\begin{subfigure}[t]{.15\textwidth}
\centering
\includegraphics[width=\textwidth]{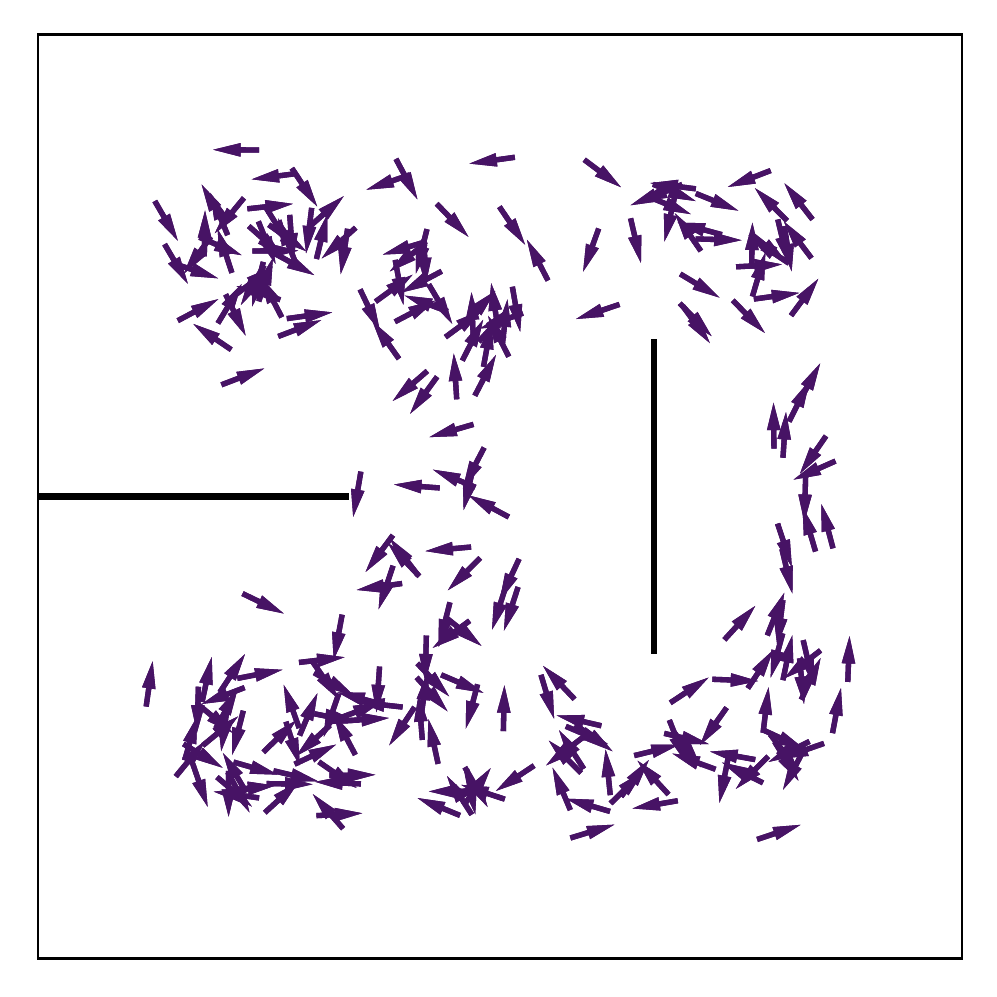}
\end{subfigure}
\hspace{\stretch{1}}
\begin{subfigure}[t]{.15\textwidth}
\centering
\includegraphics[width=\textwidth]{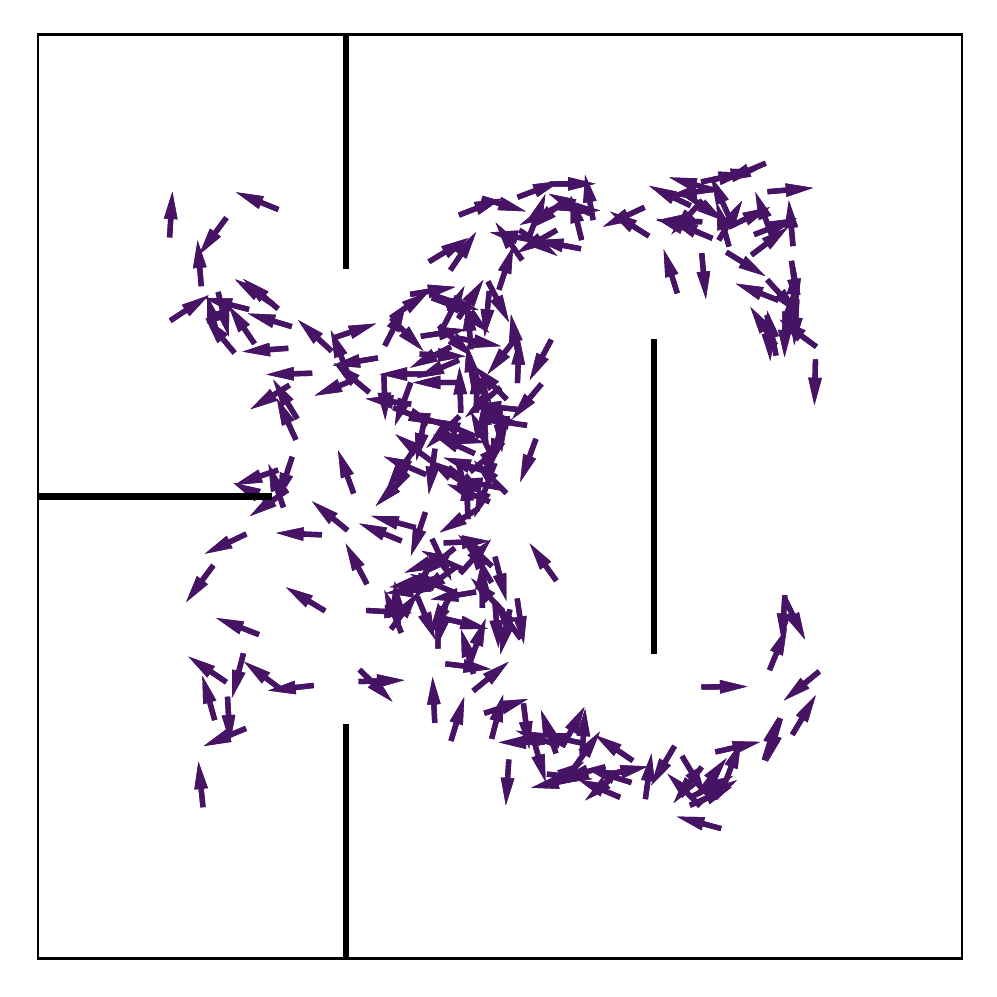}
\end{subfigure}

\begin{subfigure}[t]{.15\textwidth}
\centering
\includegraphics[width=\textwidth]{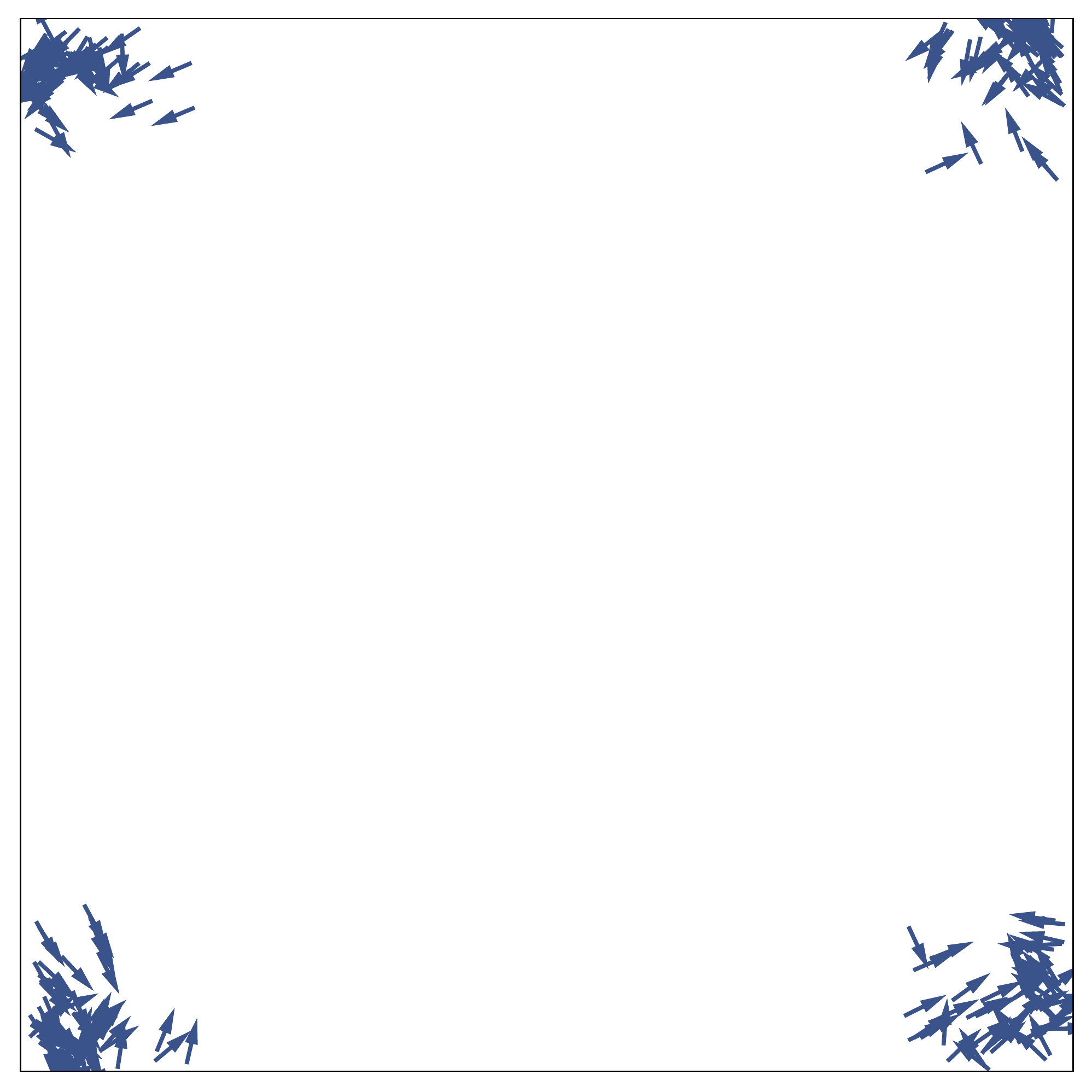}
\end{subfigure}
\hspace{\stretch{1}}
\begin{subfigure}[t]{.15\textwidth}
\centering
\includegraphics[width=\textwidth]{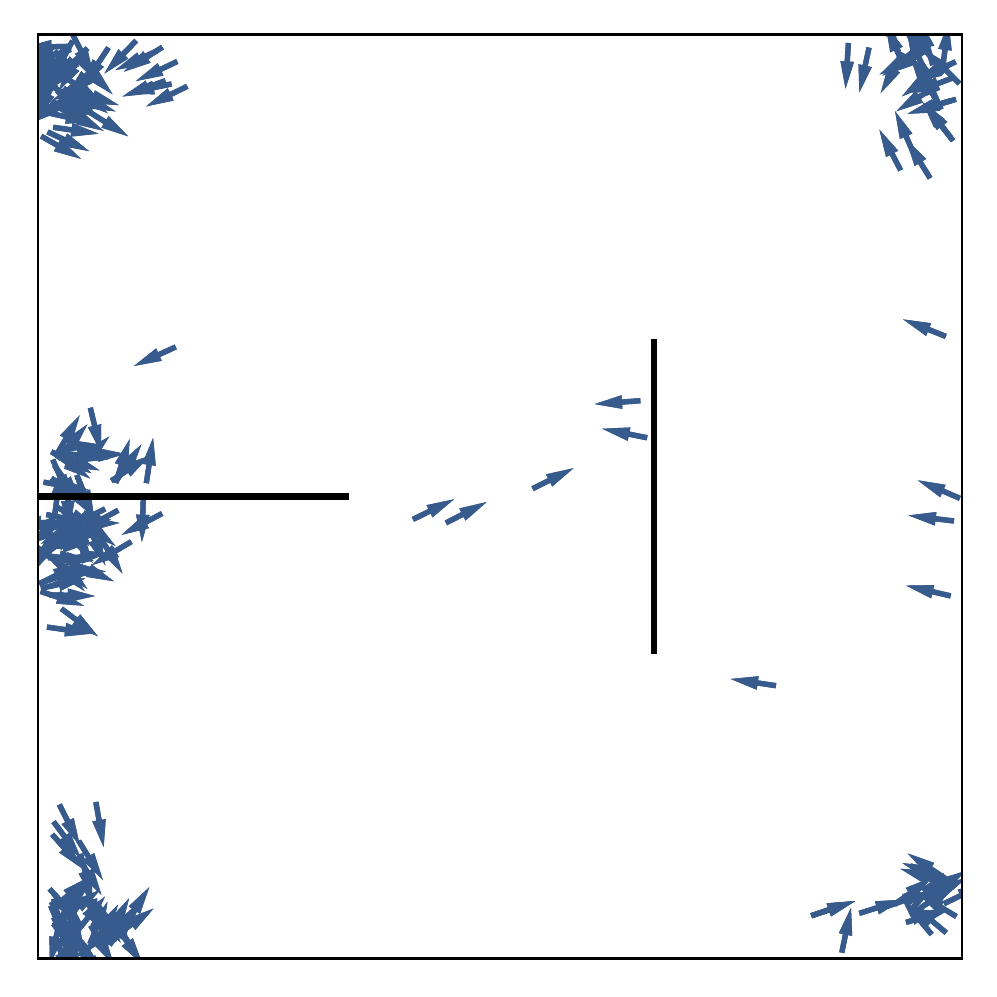}
\end{subfigure}
\hspace{\stretch{1}}
\begin{subfigure}[t]{.15\textwidth}
\centering
\includegraphics[width=\textwidth]{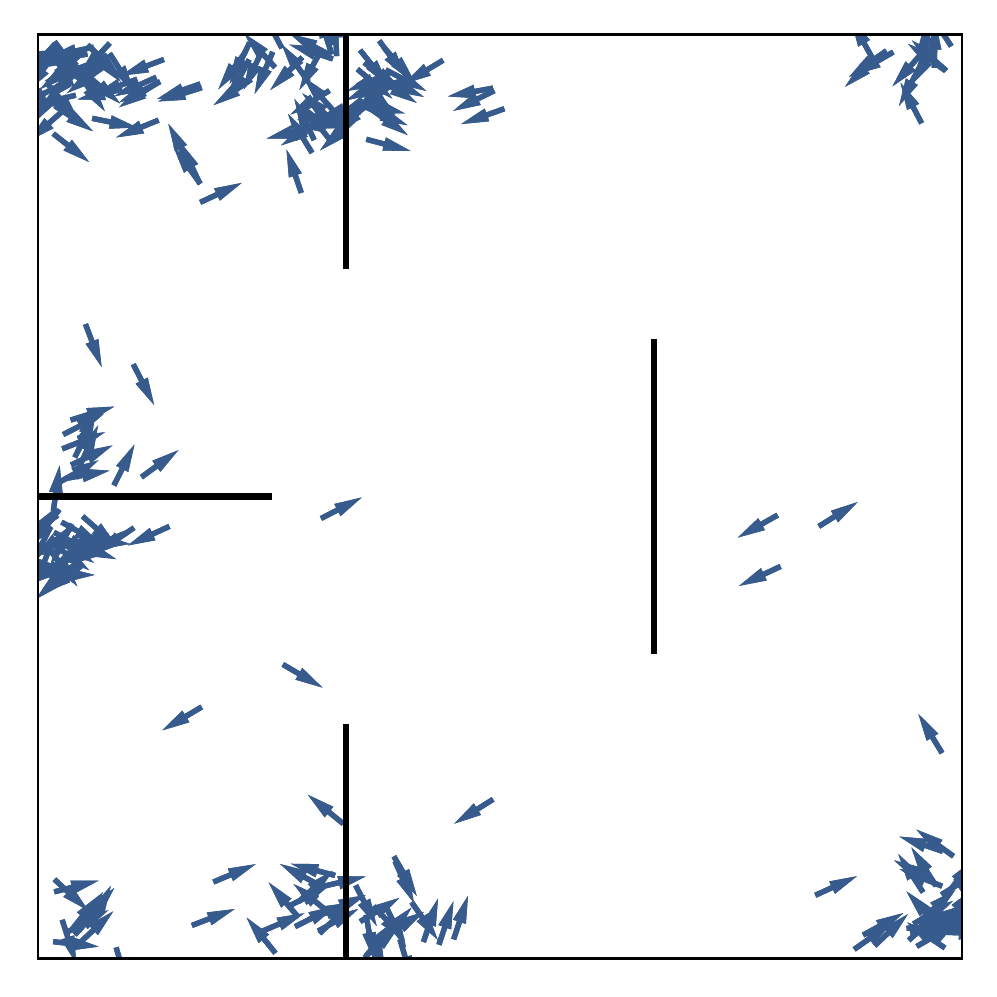}
\end{subfigure}

\begin{subfigure}[t]{.15\textwidth}
\centering
\includegraphics[width=\textwidth]{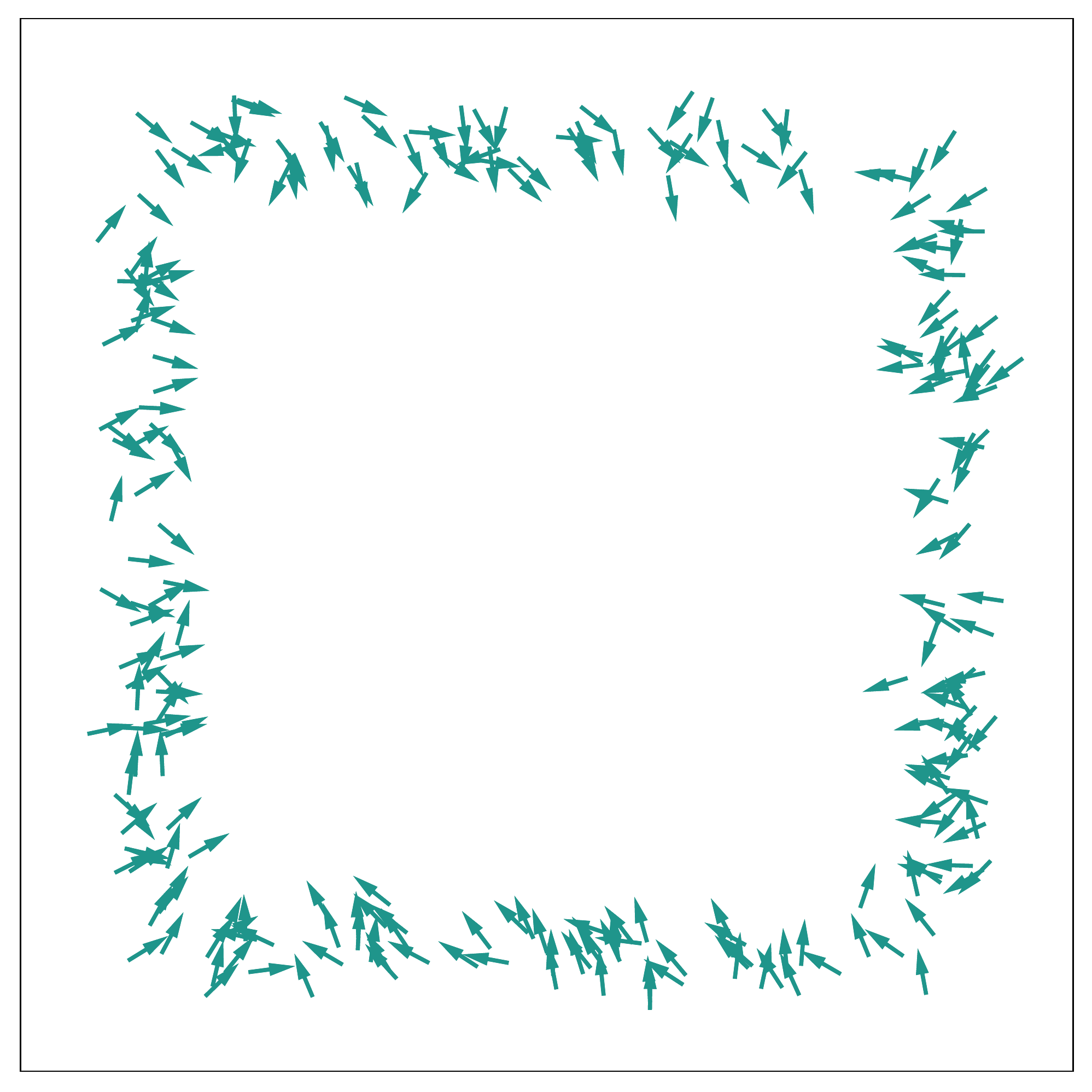}
\caption{\centering Clusters in Square}
\label{fig:square_cl12}
\end{subfigure}
\hspace{\stretch{1}}
\begin{subfigure}[t]{.15\textwidth}
\includegraphics[width=\textwidth]{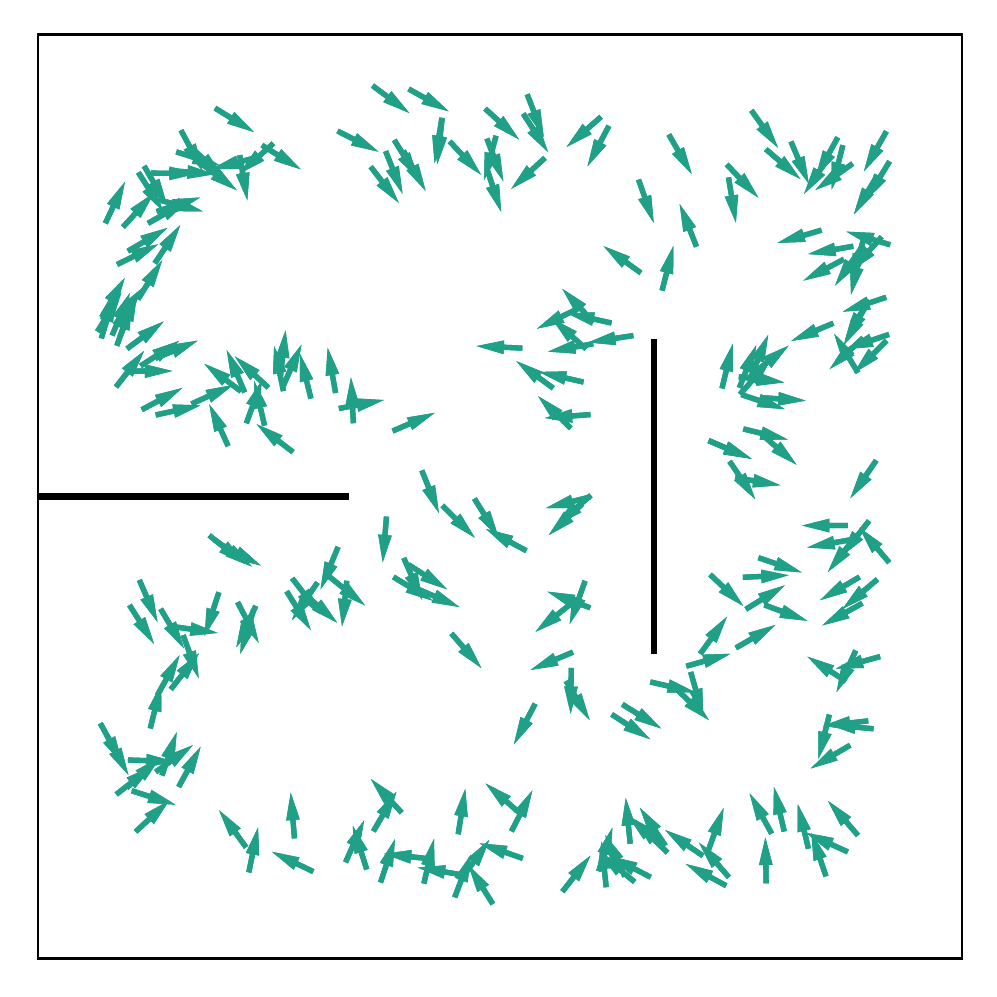}
\caption{\centering Transferred to Rooms1}
\label{fig:env2_cl12}
\end{subfigure}
\hspace{\stretch{1}}
\begin{subfigure}[t]{.15\textwidth}
\centering
\includegraphics[width=\textwidth]{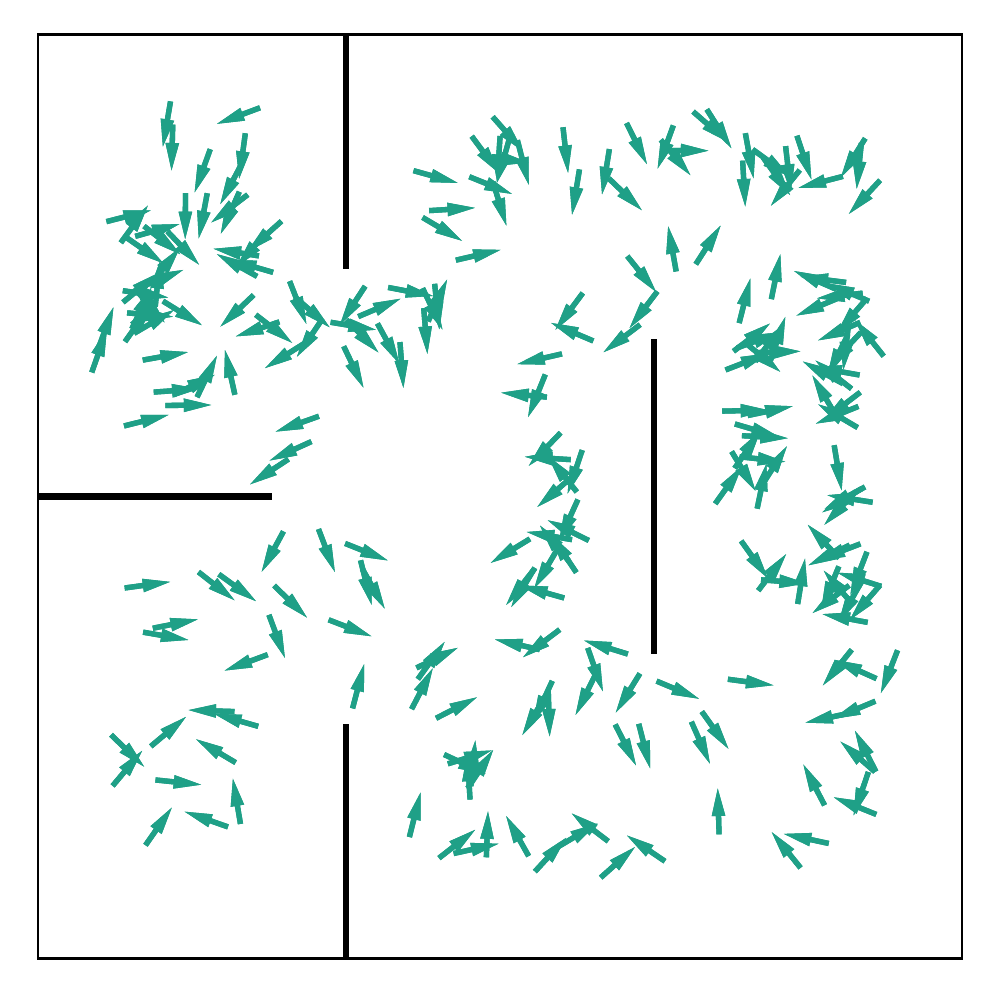}
\caption{\centering Transferred to Rooms2}
\label{fig:env3_cl12}
\end{subfigure}

\caption{Transferring some Square clusters}
\label{fig:knowledgetransfer_square}
\end{figure}

\begin{figure}

\begin{minipage}[t]{.15\textwidth}
\includegraphics[width=\textwidth]{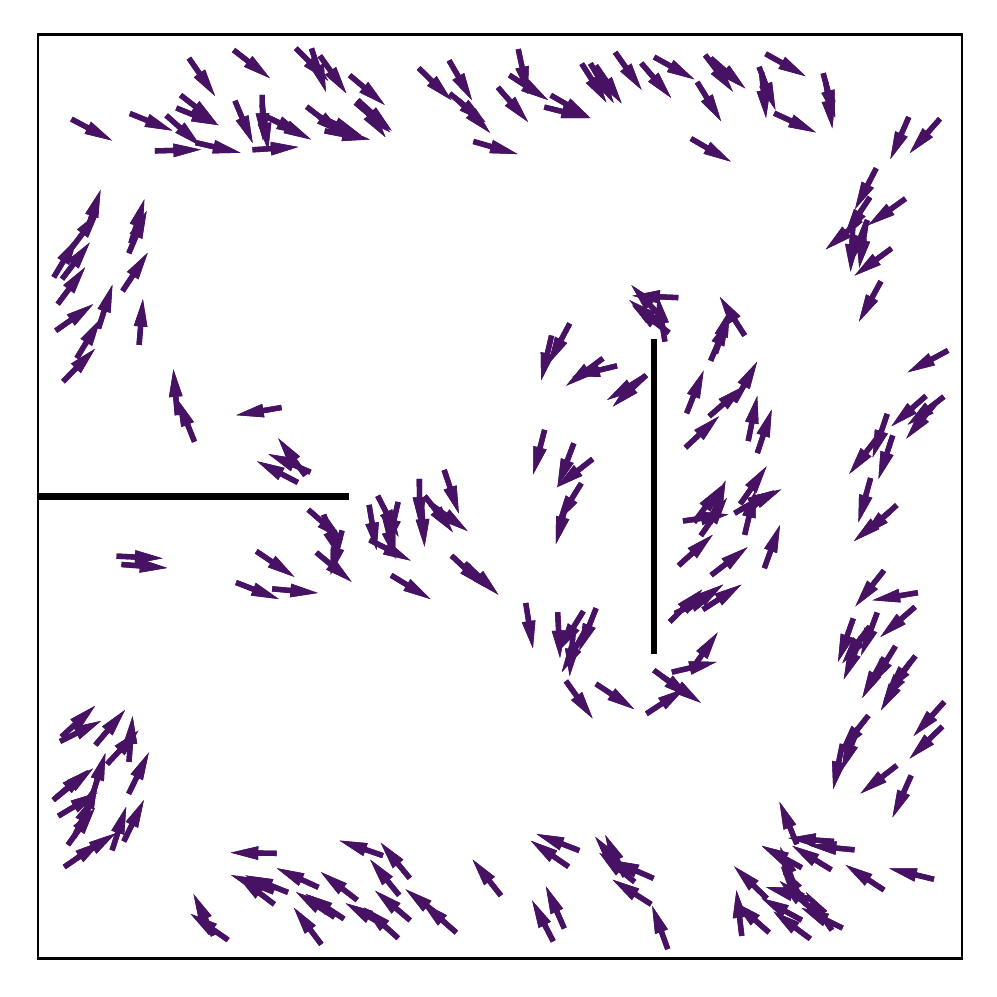}
\end{minipage}
\hfill
\begin{minipage}[t]{.15\textwidth}
\includegraphics[width=\textwidth]{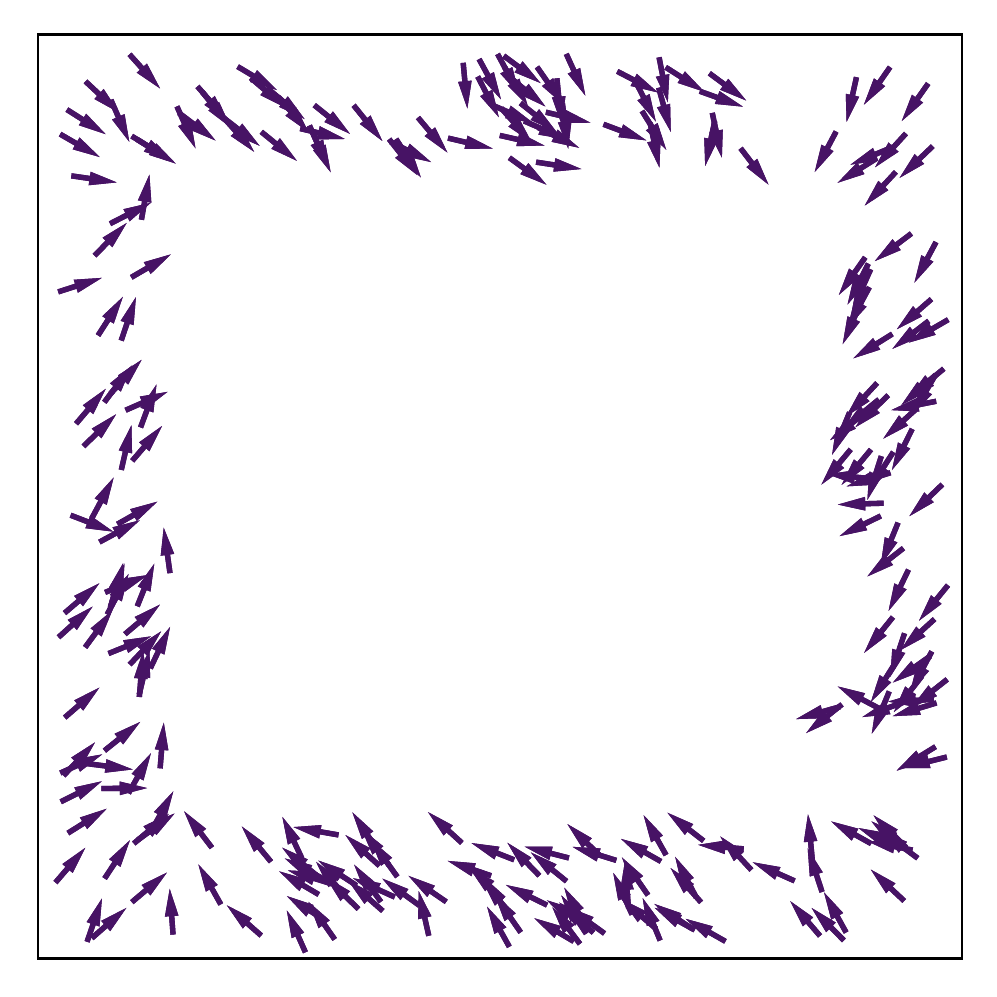}
\end{minipage}
\hfill
\begin{minipage}[t]{.15\textwidth}
\includegraphics[width=\textwidth]{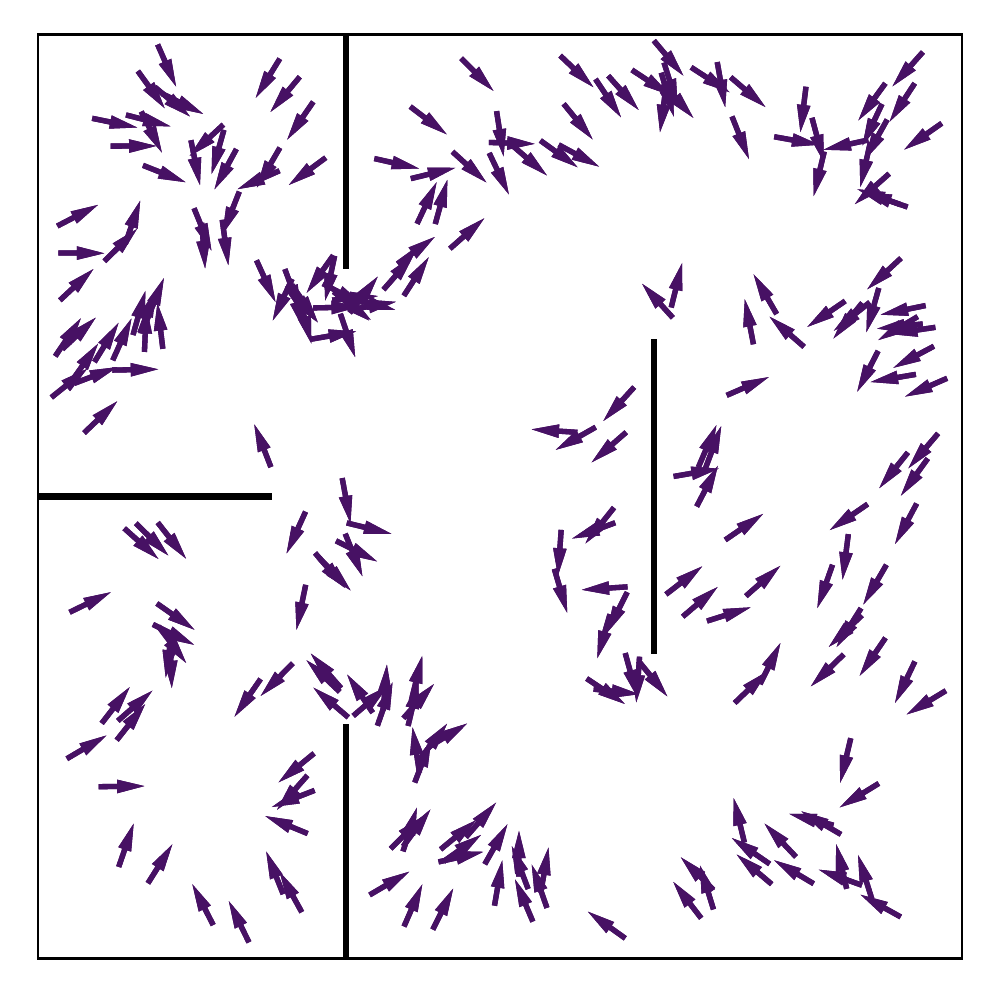}
\end{minipage}

\begin{subfigure}[t]{.15\textwidth}
\centering
\includegraphics[width=\textwidth]{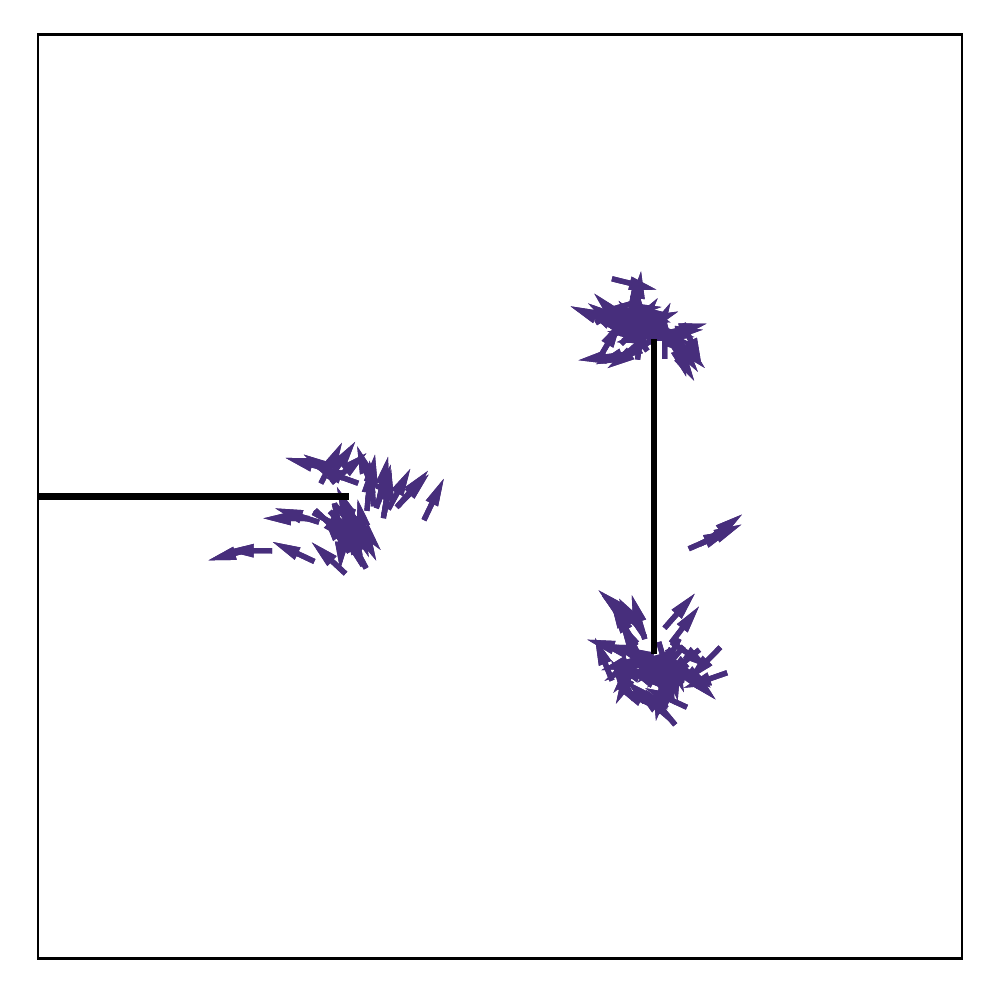}
\end{subfigure}
\hspace{\stretch{1}}
\begin{subfigure}[t]{.15\textwidth}
\centering
\includegraphics[width=\textwidth]{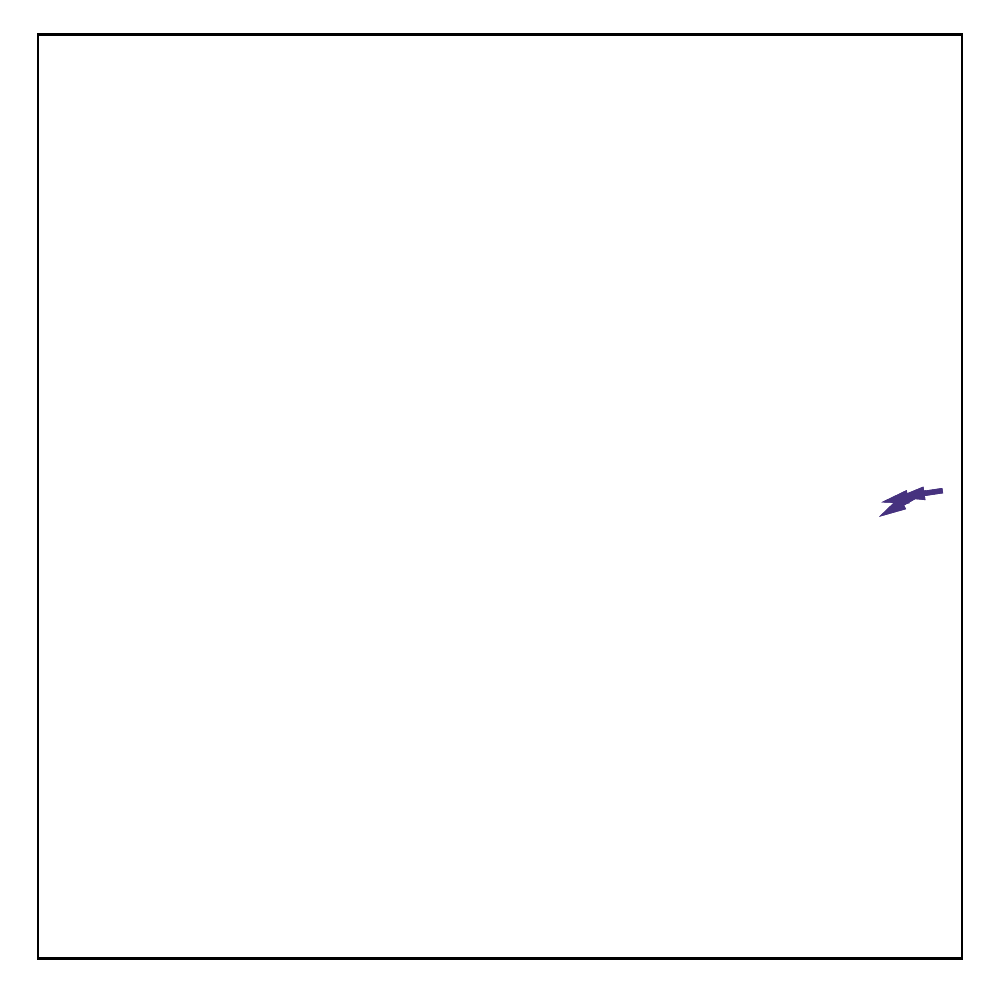}
\end{subfigure}
\hspace{\stretch{1}}
\begin{subfigure}[t]{.15\textwidth}
\centering
\includegraphics[width=\textwidth]{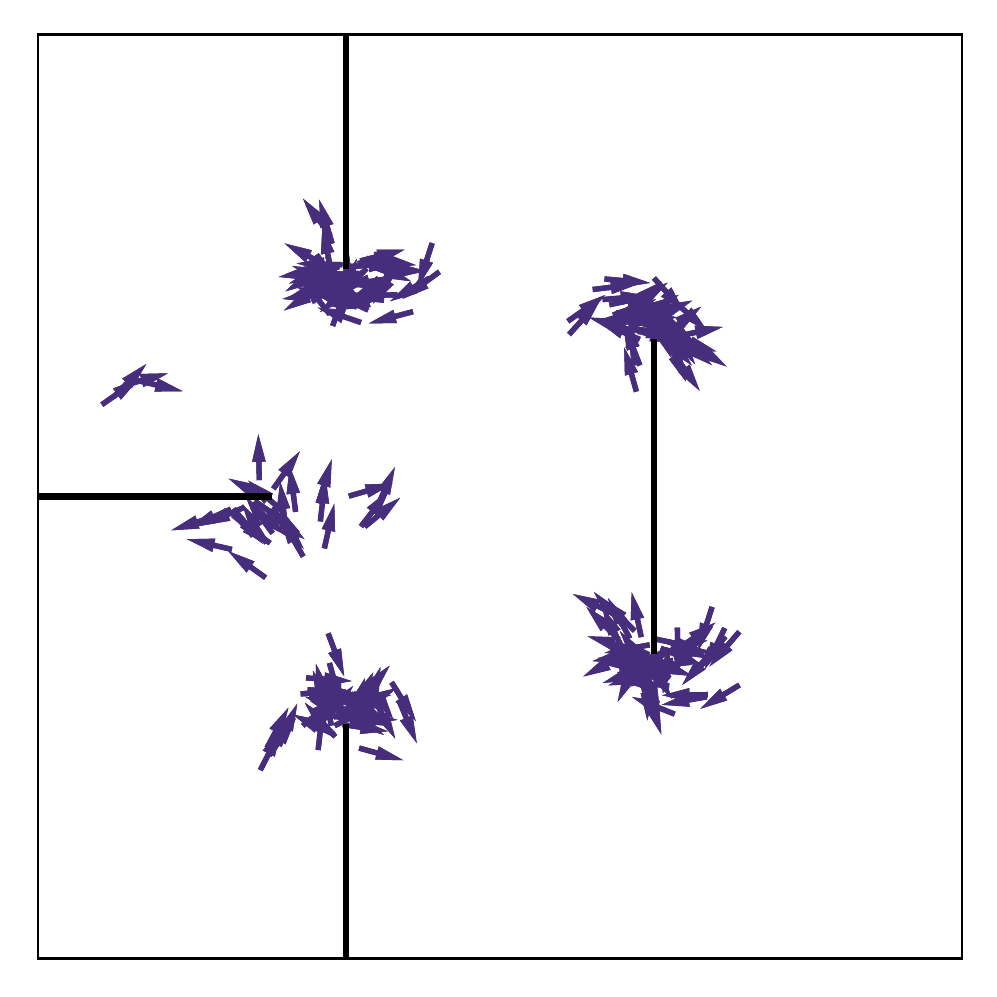}
\end{subfigure}

\begin{subfigure}[t]{.15\textwidth}
\centering
\includegraphics[width=\textwidth]{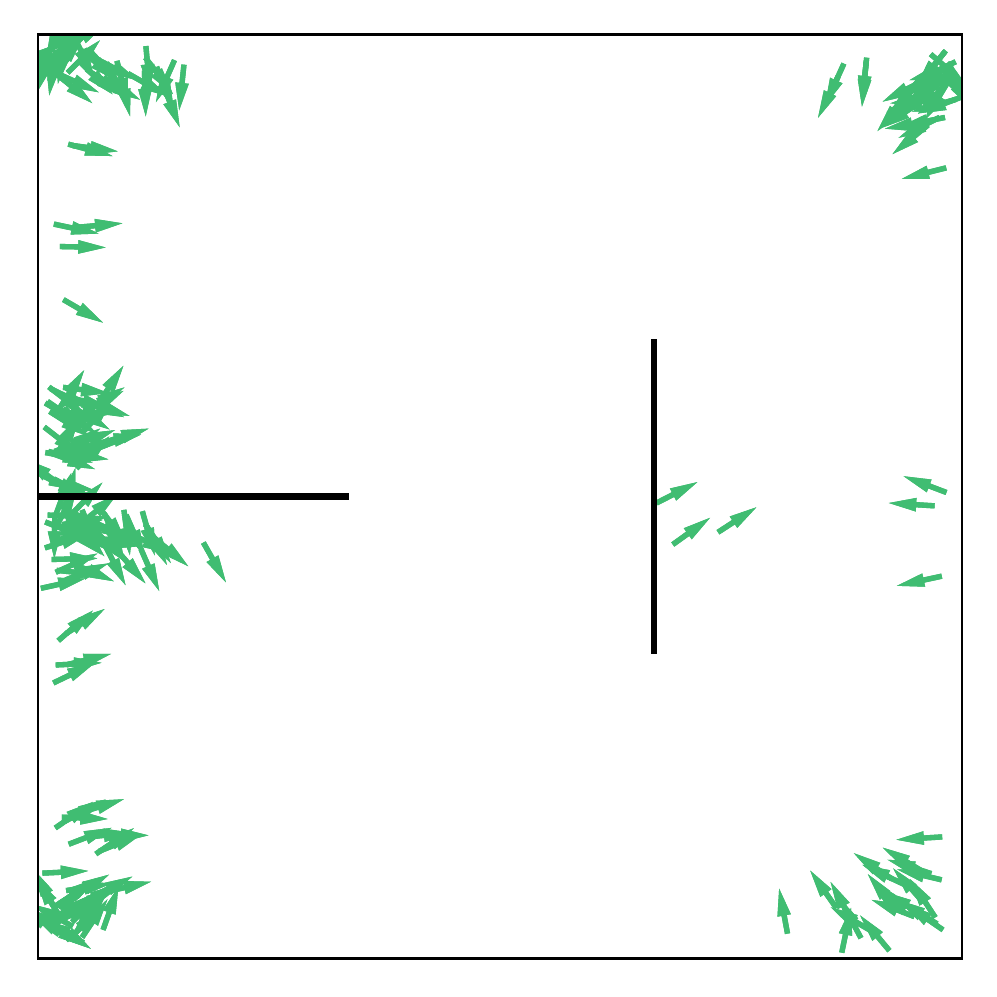}
\caption{\centering Clusters in Rooms1}
\label{fig:square_cl12}
\end{subfigure}
\hspace{\stretch{1}}
\begin{subfigure}[t]{.15\textwidth}
\centering
\includegraphics[width=\textwidth]{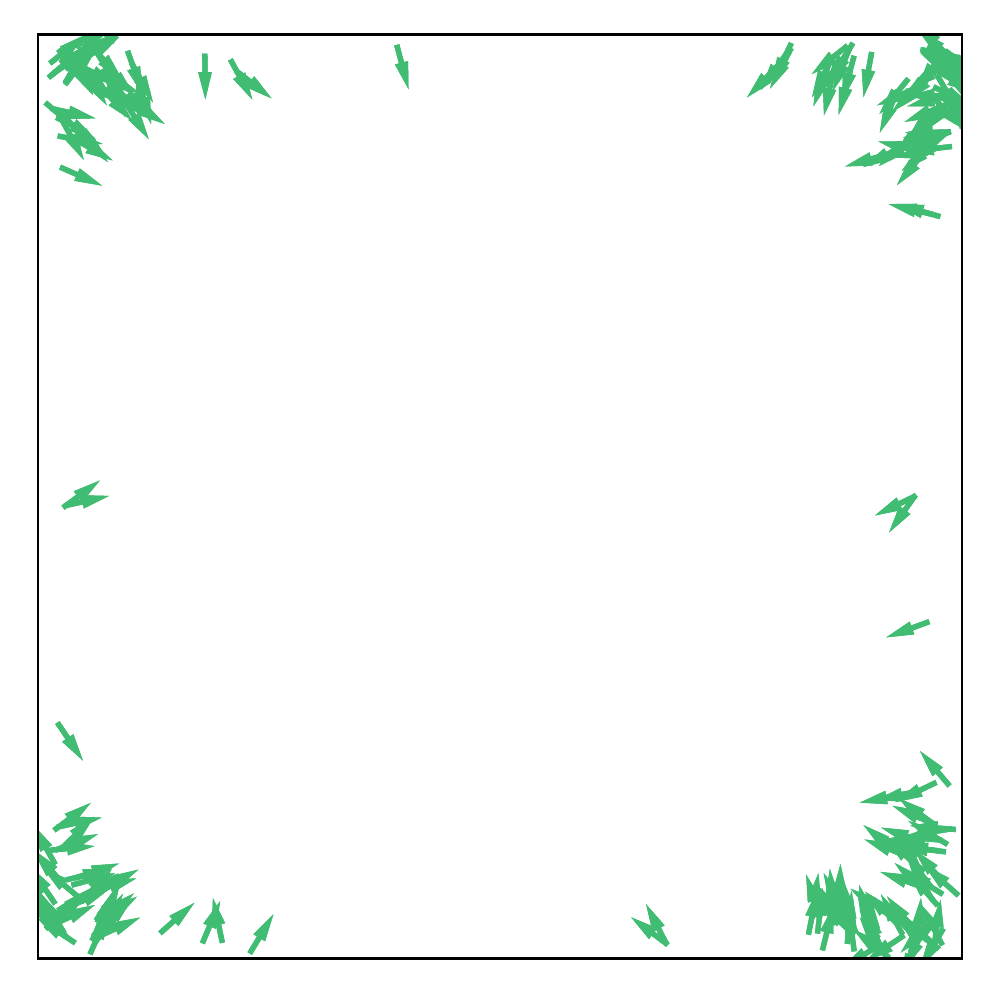}
\caption{\centering Transferred to Square}
\label{fig:env2_cl12}
\end{subfigure}
\hspace{\stretch{1}}
\begin{subfigure}[t]{.15\textwidth}
\centering
\includegraphics[width=\textwidth]{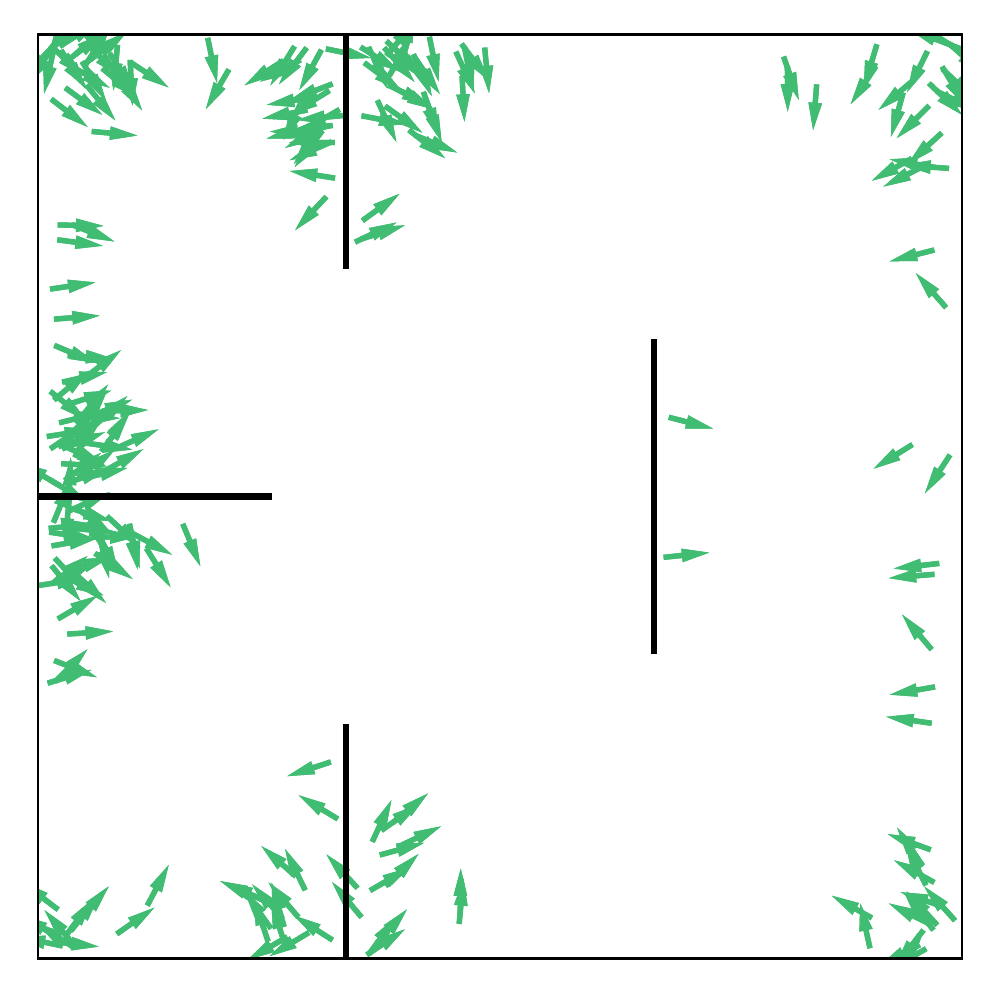}
\caption{\centering Transferred to Rooms2}
\label{fig:env3_cl12}
\end{subfigure}

\caption{Transferring some Rooms1 clusters}
\label{fig:knowledgetransfer_env2}
\end{figure}

In this experiment, we evaluate if the representations learned in one environment transfer to other environments. We train our Recurrent-SM-encoder as well as our clustering algorithm on one environment, 
then apply the learned representations and clusters in other environments. We show the transfer of some clusters learned on Square on Fig.\ref{fig:knowledgetransfer_square}. Note that the memory clusters and the corner cluster transfer well to other environments, and adapts to the new configuration of the walls. We show on Fig.\ref{fig:knowledgetransfer_env2} the transfer of a few clusters learned on Rooms1 to other environments. We observe that the knowledge is transferred well across environments. We see that the end-of-wall cluster is not transferred to square, as expected.

This shows that our model learns representations that are independent of the environment layout. This also means that the memory of the LSTM only captures short-term information, which allows the model to transfer to environments
with similar objects (wall, corners, ends of walls), but different spatial layouts. 

\section{Conclusion}

In this paper we proposed an unsupervised learning method based on sensorimotor prediction. This method allows an agent to acquire sensory representations by integrating sensorimotor information using recurrent neural networks. 

We observed that our model is indeed better at predicting the future values of sensors compared to the proposed baselines. It extracts classes of interaction with the environment that seem qualitatively more meaningful, and which contain temporal information through short-term memory of previous experiences. In particular we saw that the motor commands are very beneficial to learn representation through prediction. We also note that the cells extracted from clustering the sensory representation are similar to particular cells observed in mammals, such as distance, orientation, and border cells. More importantly, we noticed that the representation learned on an environment transfers well to others. Namely, our model doesn't remember the whole  environment but learns local sensorimotor transitions.

The approach we proposed is generic, and inspired from recent proposals about the nature and emergence of autonomy and intelligence through sensorimotor prediction (\cite{friston2010free}). It uses only raw data, and requires (in our limited experiment) very few engineering biases. In future works we want to investigate whether our approach scales to more complex environment and sensory streams. Additionally, we plan to apply the presented approach in an experiment on real robotic platforms, and in a real human environment.

One interesting possible extension would be to use the cells extracted from the sensory representation in order to learn a map of the environment. We plan to investigate how to build a graph where the nodes would correspond to activations of those cells, and the edges would correspond to motor commands. Also, we want to study the compression of such a graph of transition to obtain compact spatial representations.

In general, the proposed approach deals with very low level processing of sensorimotor streams in order to build meaningful representations. The usefulness of these representations, and how they can integrate in a cognitive architecture, would have to be demonstrated. We plan to use the learned representations in a Reinforcement Learning task. On the one hand, the success rate at the task gives a clear quantitative evaluation. On the other hand, it will allow us to evaluate the benefits of learning representations in terms of generalization, abstraction, and transfer of knowledge across different environments. 


\bibliographystyle{IEEEtran}
\bibliography{sample-bibliography}

\end{document}